\documentclass[10pt,journal,compsoc]{IEEEtran}

\ifCLASSOPTIONcompsoc
    \usepackage{cite}
\else
  \usepackage{cite}
\fi

\usepackage{subcaption}
\usepackage{graphicx}
\usepackage{booktabs}
\usepackage{flushend}
\usepackage{multirow}
\usepackage{amssymb}
\usepackage{wrapfig}
\usepackage{bm}
\usepackage{amsmath}

\DeclareMathOperator*{\argmin}{arg\,min}
\usepackage{ragged2e}
\usepackage[dvipsnames]{xcolor}  
\begin{document}

\title{Graph Representation Learning Beyond Node and Homophily}


\author{
        You~Li,
        Bei~Lin, 
        Binli~Luo,
        Ning~Gui*~\IEEEmembership{Member,~IEEE,}

\thanks{You Li, Bei Lin, Binli Luo and Ning Gui are with the School of Computer Science and Engineering, Central South University, Changsha, 410083, China e-mail: youli.syvail@gmail.com, linbei@csu.edu.cn, lblhandsome@gmail.com, ninggui@csu.edu.cn}
\thanks{Ning Gui is the corresponding author,email(ninggui@csu.edu.cn)}
\thanks{Manuscript received June 30, 2021}}

\IEEEtitleabstractindextext{
\begin{abstract}
\justifying
Unsupervised graph representation learning aims to distill various graph information into a  downstream task-agnostic dense vector embedding. However, existing graph representation learning approaches are designed mainly under the node homophily assumption: connected nodes tend to have similar labels and optimize performance on node-centric downstream tasks. Their design is apparently against the task-agnostic principle and generally suffers poor performance in tasks, e.g., edge classification, that demands feature signals beyond the node-view and homophily assumption. To condense different feature signals into the embeddings, this paper proposes PairE, a novel unsupervised graph embedding method using two paired nodes as the basic unit of embedding to retain the high-frequency signals between nodes to support node-related and edge-related tasks. Accordingly, a multi-self-supervised autoencoder is designed to fulfill two pretext tasks: one retains the high-frequency signal better, and another enhances the representation of commonality. Our extensive experiments on a diversity of benchmark datasets clearly show that PairE outperforms the unsupervised state-of-the-art baselines, with up to  101.1\% relative improvement on the edge classification tasks that rely on both the high and low-frequency signals in the pair and up to 82.5\% relative performance gain on the node classification tasks. 

\end{abstract}
}

\maketitle

\section{Introduction}
Graph structured data has become ubiquitous in various practical applications, such as molecular, social, biological, financial~\cite{Zhang18survey} and other networks.  Due to its high dimensional data and complex structure, one of the primary challenges in graph representation learning is to find a way to represent, or encode, graph structure so that it can be easily exploited by machine learning models\cite{Hamilton17_review}. Unsupervised Graph Representation Learning(GRL) distills various graph information into a dense vector embedding in a downstream task-agnostic way. We believe that it should contain all the information about the graph, including low-frequency signals from neighbors in the same classes and high-frequency signals from neighbors in different classes~\cite{bo21}. 


However, most existing graph representation learning methods assume that the geometric closed nodes are more similar. Under this assumption, different streams of GRL algorithms, including both random walk-based and GNN-based solutions, are proposed so that similarity in the embedding space approximates similarity in the original network\cite{Grover16, hou2019representation}. For the stream of random walk-based approaches, its fundamental design goal is to optimize the node embeddings to match their co-occurrence rate on short random walks over the graph~\cite{Hamilton17_review}. Most existing GNNs usually exploit the low-frequency signals of node features, e.g., GCN~\cite{Kipf16} and GAT~\cite{Veli17_GAT}, and can be seen as a particular form of the low-pass filter from the node perspectives~\cite{Nt19,xu-heat}. The low-pass filter is of great importance for their success in the significant downstream tasks\cite{wu2019simplifying}, e.g., node classification and link prediction, since many datasets are indeed homophily networks\footnote{https://en.wikipedia.org/wiki/Network\_homophily}, i.e., nodes with similar labels and features tend to connect. However, many real-world networks are heterophily networks, i.e., nodes from different classes tend to connect. For example, the chemical interactions in proteins often occur between different types of amino acids\cite{bo21}. Even for homophily networks, the actual level of homophily may vary within different pairs of node classes\cite{Zhu2020}. This reminds us that the similarity assumption used by existing GRL solutions is distant from optimal for real-world scenarios. Not all the downstream tasks benefit from the akin embeddings between connected nodes.


\begin{figure}
  \centering
  \centerline{\includegraphics[width=.95\columnwidth]{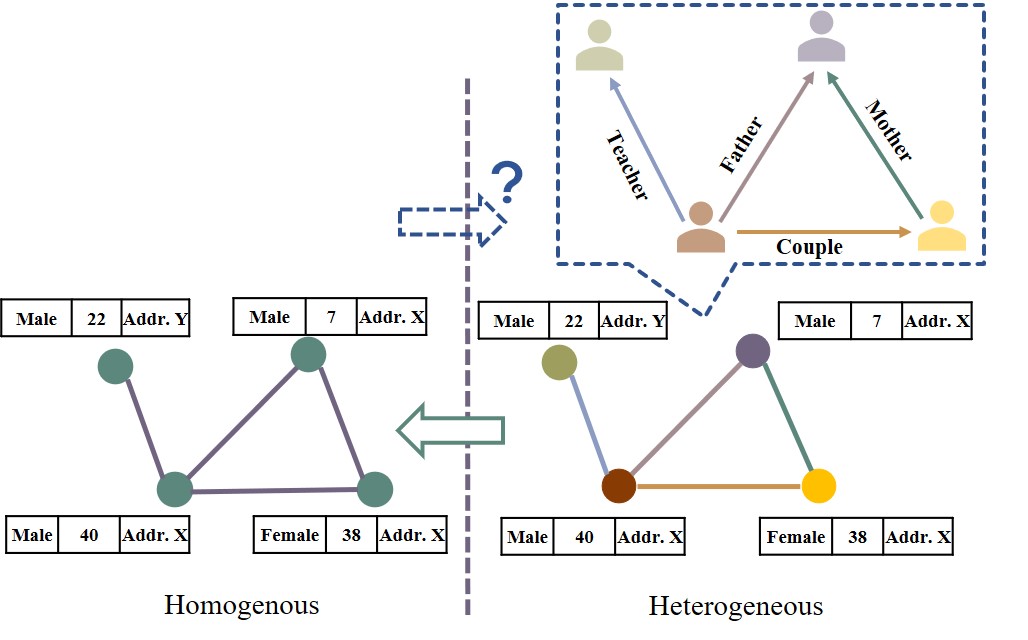}}
  \caption{Inferable relations based on pairs' feature patterns}
  \label{fig:view}
\end{figure}

The edge classification task, for instance, demands the reservation of feature differences between paired nodes. As shown on the right side of Fig.~\ref{fig:view}, in a multi-layer social network, a node representing a person may have complex relations with other nodes, which is embodied in different types of edges. Although we often only have a desensitized homogeneous graph(left part) of the original graph without any relation information, it is still possible to infer related implicit relationships through the combination of both low and high-frequency signals among features. For instance, an edge between two nodes with the same address(low-frequency signal) is more likely to have the "live with" relation. The difference in age with the same address might also be a good hint of a father/mother-son relationship. Thus, in the domain of semi-supervised learning, several recent approaches \cite{abu2019mixhop,Hou20Measuring} realize the challenges of learning from graphs with high-frequency signals and propose a series of designs: e.g., low and high-frequency filters~\cite{Zhu21,bo21}, and the separation and combination of ego and neighbor embedding training~\cite{Zhu2020}, to facilitate the learning in heterophily networks. However, these approaches limit the discussion in the node homophily and rely on the node labels for learning. Therefore, with only graph structure, node features, and no labels, effectively distill different information and make the learned representation genuine task-agnostic remains largely unexplored. Capabilities in translating both low and high-frequency signals during embeddings are essential to support diverse downstream tasks.

Since many existing graph representation learning models either assume strong homophily and fail to generalize to networks with heterophily or rely on the guidance of labels and cannot effectively adapt to unknown downstream tasks. To break these limitations, we propose PairE, a novel multi-task unsupervised model to preserve both information for homophily and heterophily by going beyond the localized node view and utilizing higher-level entities with more prosperous expression powers to represent feature signals beyond homophily. The main contributions are summarized as follows:

\noindent\textbf{Current Limitations: } In the domain of unsupervised GRLs, we first reveal the limitation that exists on the node similarity assumption as their evaluation is limited to a few benchmarks with similar properties. We further point out that they suffer significant performance degradation when downstream tasks demand different entities' low and high-frequency signals.

\noindent\textbf{Key Designs:} We propose a set of crucial designs to boost the integration of high-frequency signals without trading off accuracy for tasks demanding homophily. (D1) Pair-view embeddings:  one natural structure - Pair - existing in the graph is used as the basic unit for embedding. The high-level entity retains both the commonality and difference of pair features simultaneously and avoids over-smoothing the pair features. (D2) A novel multi-task self-supervised autoencoder integrates both the low-frequency and high-frequency information of a pair and its neighbors into the latent vectors with two different reconstruction pre-tasks. (D3) "Skip Connection" and Concatenation to preserve the integrity and personalization of different inputs.

\noindent\textbf{Outstanding experimental results: } Comprehensive experimental results show that compared to unsupervised approaches, PairE has unprecedented performance in a variety of downstream tasks: up to 101.1\% improvement on the multi-label edge classification tasks, and up to 82.5\% performance gain on the node classification task. 

Source code is available at\footnote{https://github.com/syvail/PairE-Graph-Representation-Learning-Beyond-Node-and-Homophily}.

\section{Related Work}
\noindent\textbf{Learning under homophily.} Early unsupervised methods for learning node representation mainly focus on matrix factorization methods, which calculate losses with handcrafted similarity metrics to build vector representations for each node\cite{Wang19-seed} with latent features. Later, TADW\cite{Yang15,yang2019low}, etc. are proposed the integration of node features into the embedding process, which does not pay attention to the information of different frequencies in the node features. The inspiration for random walks comes from the effective NLP method. The difference between these methods lies in the strategy of walking(e.g., DFS, and BFS)\cite{Perozzi14, Tang15, Grover16}. A few recent approaches notice the need to encode higher-level of graph objects, e.g., SEED encodings lines of walks\cite{Wang19-seed}. ProGAN~\cite{gao2019progan} encodes the similarity relationship between different nodes by generating the proximity between nodes. Edge2vec~\cite{wang2020edge2vec} proposes the explicit use of the edge as basic embeddings units. However, they are based on the assumption of neighborhood similarity in the graph topology and are hard to integrate node features.


Many GNN-based solutions typically adopt the so-called message mechanism to collect neighboring information with general differences on how to design the function of aggregation, combine, and readout functions~\cite{Xu18_gin}. H2GCN\cite{Zhu2020} points out; many GNN solutions are optimized for network homophily by propagating features and aggregating them within various graph neighborhoods via different mechanisms (e.g., averaging, LSTM, self-attention)~\cite{Kipf16,hamilton_grapshsage, Veli17_GAT, hou2019representation}. GraphHeat~\cite{xu2020graph} designs a more powerful low-pass filter through heat kernel. Another stream is based on autoencoder, e.g., GAE and VGAE~\cite{GAE16} and their follow-up work ARVGE~\cite {ARGA18} with an adversarial regularization framework, CAN~\cite{meng2019co} embeds
each node and attribute with a variational auto-encoder. Several approaches try to keep more complex graph information by learning with higher-level entities, e.g., K-GNN~\cite{123gnn19} learns the representation by performing hierarchical pooling combinations from nodes under semi-supervised settings. However, those solutions are also based on the assumption that connected nodes should be more similar. Therefore, they fail to generalize to networks with heterophily.

\noindent\textbf{Semi-supervised GNN for heterophily.} Several recent works have noticed the importance of modeling both the homophily and heterophily of graphs. CS-GNN~\cite{Hou20Measuring} has investigated GNN’s ability to capture graph useful information by proposing two diagnostic measurements: feature smoothness and label smoothness, to guide the learning process. To better preserve both high-frequency and low-frequency information, H2GCN~\cite{Zhu2020} separates the raw features and aggregated features during the learning processes. FAGCN~\cite{Zhu21,bo21} designs two novel deep GNNs with flexible low-frequency and high-frequency filters, which can well alleviate over-smoothing. Geom-GCN~\cite{pei2020geom} utilizes the structural similarity to capture the long-range dependencies in heterophily networks. However, the discussion of their work is limited to the discussions of the node-view heterophily. Furthermore, they are semi-supervised and rely on the node labels to guide the learning processes. In the settings of unsupervised GRLs, how to effectively encode different frequency information of different entities into embeddings and make the graph representation can effectively support a wealth of unknown downstream tasks remains largely unexplored.  

\noindent\textbf{Learning for relations.} In the domain of multi-layer networks, TransE~\cite{TransE13} learns vector representations for both entities and relations by inducing a translation function to represent the relations. Later works~\cite{TransR15,Rotate19} extend this concept with different relations with different score functions. PairRE~\cite{PairRE} further proposes a model with paired vectors for each relation representation. This approach also adopts the paired vectors for paired nodes. These solutions all explicitly demand the labels of edges, which are not available in the settings of GRL.


\section{Notations and Preliminaries}

\subsection{Notations and definitions}

There are two natural entities in the graph: nodes and edges. Let $G = (V;E;X)$ denote a directed graph, where $V$ is the node set with $n$ nodes and $E$ is the edge set, with node feature matrix $X \in \mathbb{R}^{n \times d}$. We denote a general neighborhood centered around $u$ as ${N}(u)$ that does not include the node $u$. The $d$ dimensional feature vector of node $u$ is denoted as $x_u$.


\begin{wrapfigure}{r}{0.45\columnwidth}
\vspace{-0.25in}
\centering
\includegraphics[width=0.45\columnwidth]{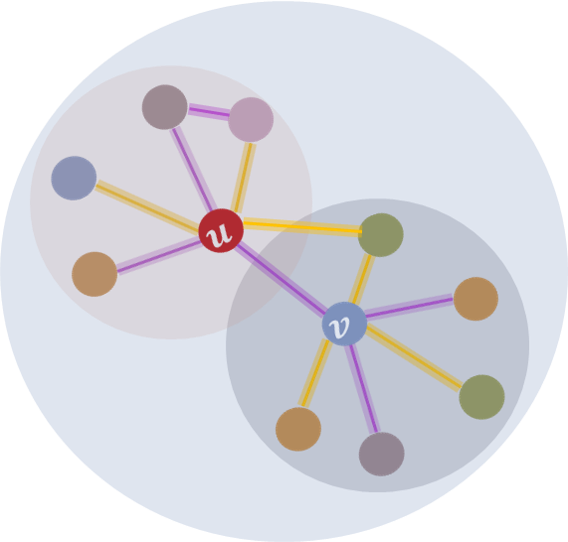}
\caption{The Node \& Pair View}
\label{fig:nodepair}
\vspace{-0.2in}
\end{wrapfigure}



\noindent \textbf{Definition. 1}: (Pair) A pair is an ordered node pair $p_{u,v}$ starting from the node $u$ to the node $v$, where $u, v \in V$ and $(u,v) \in E$. Node $u$ and $v$ are called the source node and the target node, respectively. We denote $P(G)$ as the set of all pairs in $G$.

\noindent \textbf{Definition. 2}: (Pair neighborhood). Different GNN models have different definitions of neighbor: a general neighborhood centered around $p_{u,v}$ as $N(p_{u,v})$ ($G$ may have self-loops), the corresponding neighborhood that does not include the $p_{u,v}$ itself as $N(p_{u,v})$, and the general neighbors of pair $p_{u,v}$ at exactly i-hops/steps away (minimum distance) as $N_i(p_{u,v})$. In this paper, We adopt the typical definition of 1-hop neighbors, and define the neighborhood of $p_{u,v}$ as:
    
    $$N(p_{u,v})= \{p_{u,u_i} \cup p_{v,v_j}\}$$
    $$p_{u,u_i},p_{v,v_j} \in P(G), u_i \in N(u),v_j \in N(v) $$
The bigger circle in Fig.~\ref{fig:nodepair} shows the neighborhood of $p_{u,v}$, while the two smaller circles show the 1-hop neighborhood of node $u$ and $v$ respectively. 
    

  
\noindent \textbf{Definition. 3}: (Pair embedding)  The task of pair-based graph embedding is to learn a low-dimensional vector $z_{u,v}$ to represent the ordered node pair $p_{u,v}$ in the entire graph. In the embedding process, complete feature information of the node pair and surrounding neighboring information $N(p_{u,v})$ should be preserved as much as possible.

\subsection{Global/Local v.s Node/Edge label assortativity}

Here, we define the metrics to describe the essential characteristics of a graph. There exist multiple metrics, e.g., the label smoothness defined in~\cite{Hou20Measuring} or the network assortativity in~\cite{Newman2003}. We adopt assortativity as it can be easily generalized to nodes and edges. Here, the assortativity coefficient measures the level of homophily of the graph based on some vertex labeling or values assigned to vertices. If the coefficient is high, connected vertices tend to have the same labels or similar assigned values.

   
\noindent\textbf{Definition 4.} (Global Node/Edge label assortativity) For a graph $G$ with node labels, the node label assortativity is defined as follows:
\begin{equation}
\begin{aligned}
    r_n^{global} = \frac{\sum_{i}e_{ii}-\sum_{i}a_{i}b_{i}}{1-\sum_{i}a_{i}b_{i}} =  \frac{Tr(\textbf{e})-\sum(\textbf{e}^2)}{1-\sum(\textbf{e}^2)}\\
\end{aligned}
\end{equation}
    where $\textbf{e}$ is the matrix whose elements are $e_{ij}$, $e_{ij}$ is the fraction of edges connecting nodes of label i and label j, $a_i = \sum_{j}e_{ij}$, and $b_i = \sum_{i}e_{ij}$. For multi-label nodes (L-dimensional labels), we can calculate the corresponding assortacity level according to the global distribution of each dimensional label, and finally obtain the assortavity set of all labels: $\{r_n^{global}(1),...,r_n^{global}(L)\}$.

  The global edge label assortativity is denoted as  $r_e^{global}$ , which uses edges as the unit to measure the global label distribution of edge and edge neighborhoods in the entire graph. Its calculation method is similar to $r_n^{global}$. Similarly, $\{r_e^{global}(1),...,r_e^{global}(L)\}$.

When nodes with similar labels tend to connect with each other, $r^{global}_n(l) \rightarrow 1$, the graph exhibits strong homophily from the $l^{st}$ label-view of node. The graph displays node heterophily when $r^{global}_n(l) \rightarrow 0$, with dissimilar nodes connected with others. $r^{global}_e$ measures the edge homophily of the graph with similar traits.



\begin{figure}[htb]
\centering
\begin{subfigure}{0.49\columnwidth}
\centering
\includegraphics[width=0.98\columnwidth]{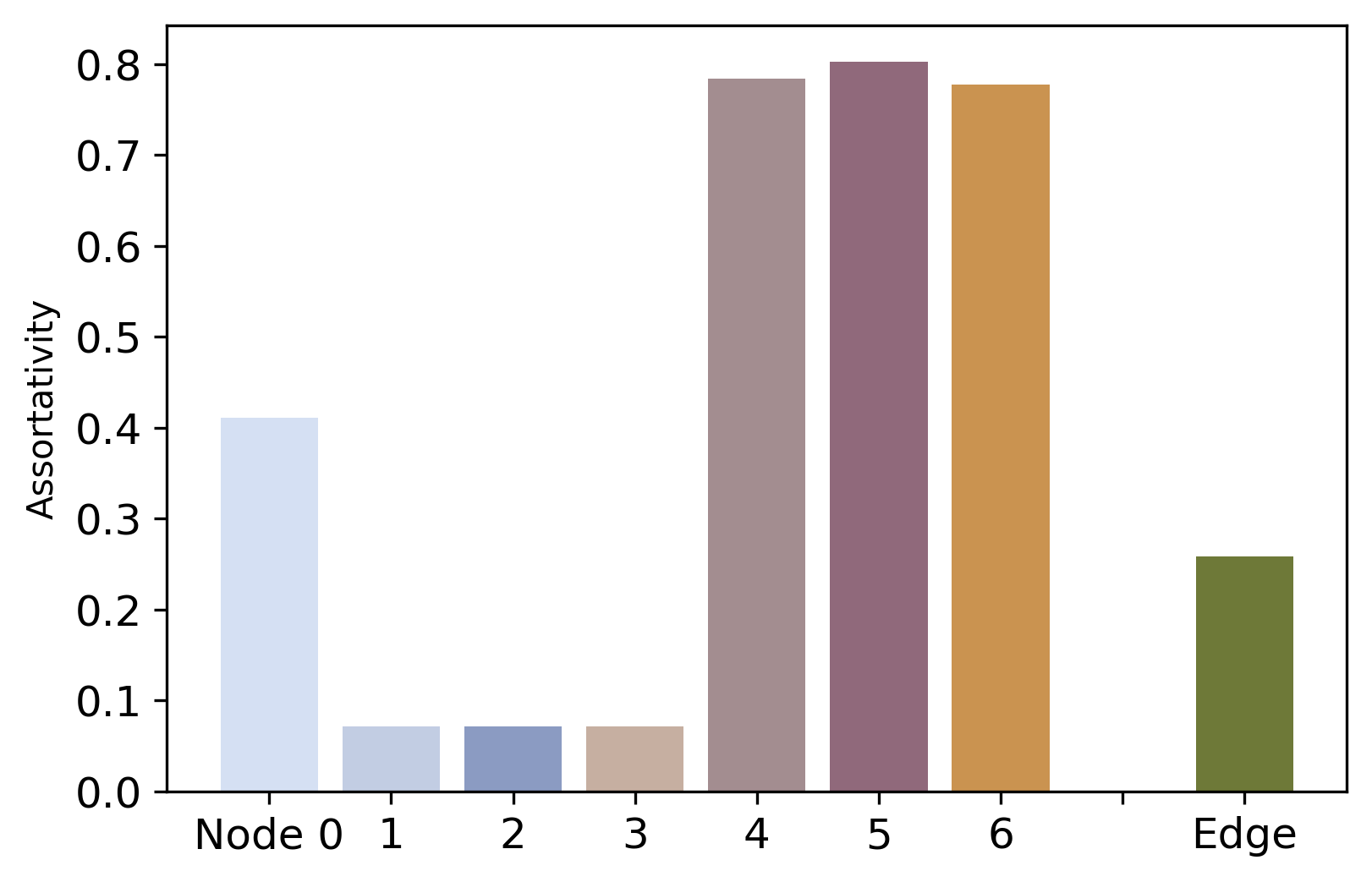}
 \caption{The global label assortativity of Cuniform}
\end{subfigure}
\begin{subfigure}{0.49\columnwidth}
\centering
\includegraphics[width=0.98\columnwidth]{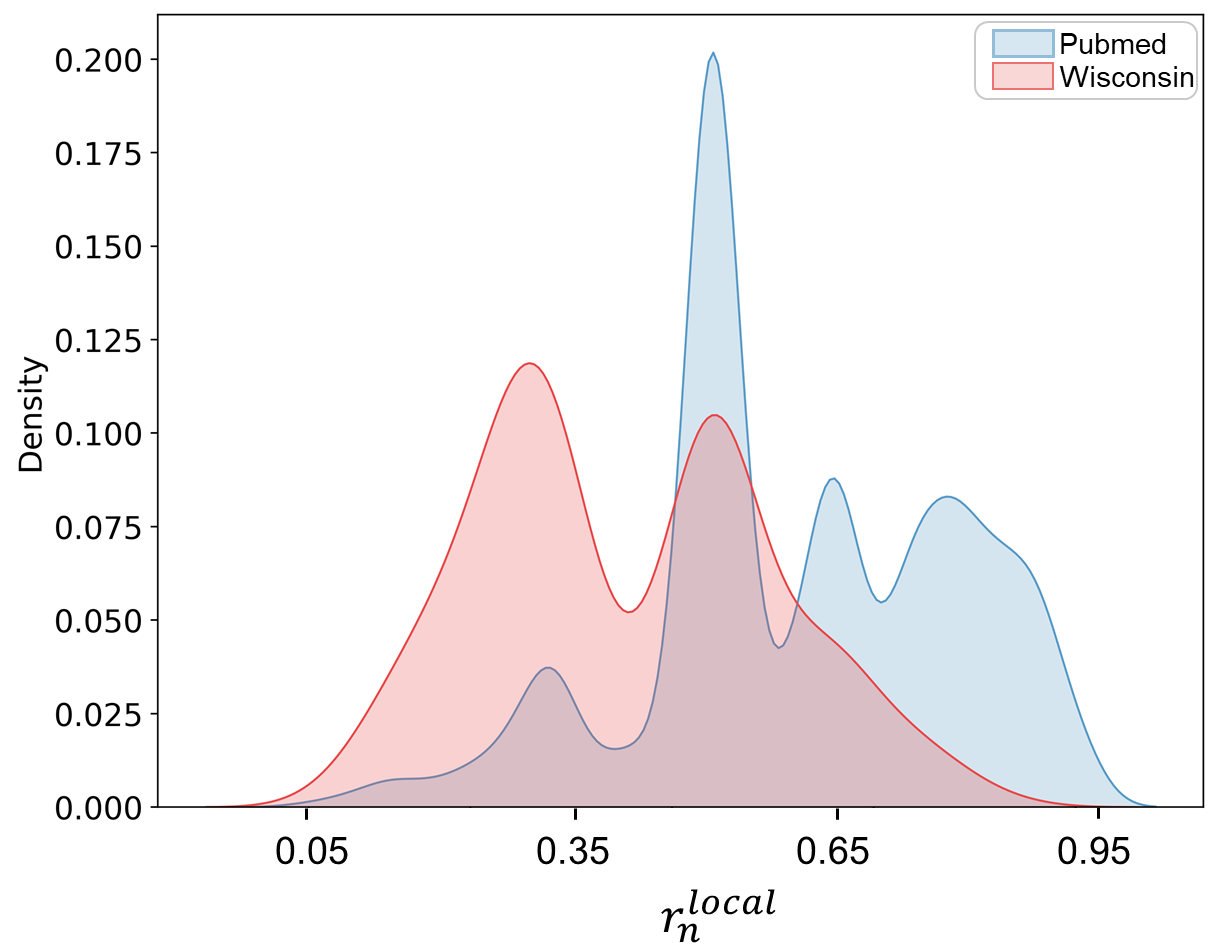}

 \caption{the local label assortativity}
\end{subfigure}
\caption{The global/local node/edge assortativity }
\label{fig:graph-assortativity}
\end{figure}
 
As mentioned in ~\cite{suresh2021breaking}, the network performance in the real world often shows diversified mixed modes: the level of assortativity of different nodes or edges is inconsistent. To better represent the diverse patterns in the network and accurately quantify and compare the learnability of pair and other benchmarks in different patterns, we adopt a generalized concept, the local node/edge-level assortativity, one metric  based on the label distribution from an individual node/edge-view. 

\noindent \textbf{Definition 5.}(Local Node/Edge label assortativity) We define the  $r_n^{local}(u)$ as a measure of the local assortativity level for node $u$:
\begin{equation}
\begin{aligned} 
    r_n^{local}(u) = \frac{|\{v\}:v\in N(u)\land y_u=y_{v}|}{|N(u)|};u \in V 
\end{aligned}
\end{equation}

where $|N(u)|$ denotes the number of neighbors of node u, $y_u$ denotes the label of node $u$. Due to concise consideration, we only provides node-label assortativity. For edge $e$, the $r_e^{local}(e)$ measure the local label distribution of edge $e$ and its neighborhoods in promximity. Its calculation method is similar to $r_n^{local}(u)$.  It is worth noting that the definition and calculation of assortativity used in this article are for class-level, not cross-class.

\subsection{Discussion}

From definition 4, we can see that one graph is a complex structure, and it can be seen from many different perspectives: node and edge. Existing approaches, however, limit their discussions on the node homophily, which is by no means the only important metric for embeddings. Furthermore, the following observations are identified:

\noindent\textbf{Observation 1.} \textit{One graph exists multiple entities and levels of homophily.}

According to the definition, the node and edge label assortativity coefficients are two different metrics to describe the graph to present the overall trends for all the node/edge labels. For many real-world networks with multiple node labels, multiple node assortativity coefficients exist. The same holds for multi-label edges. From our observation, one graph might have discrepancy highly in those assortativity coefficients. Fig.~\ref{fig:graph-assortativity} shows the node/edge assortativity for Cuniform, a graph with 7-dimension node labels and 1-dimension edge label(each dimension has two values: 0 or 1). It clearly shows that one graph might display both homophily and heterophily of nodes and edges: the assortativity of different node labels can range from $0.07$ to $0.80$. Thus, we can not simply classify a network as node-based homophily or heterophily. It is not appropriate to use the homophily assumption for GRL as it is only one one graph perspective. For different classification tasks, the information required can be totally different.


Furthermore, as pointed out by \cite{Zhu2020}, the actual level of assortativity may also vary within different pairs of node classes, i.e., there is a different tendency of connection between each pair of classes with two different characteristics of the graph. Thus, some pairs of classes may exhibit homophily within the same network, while others exhibit heterophily. In Fig.~\ref{fig:graph-assortativity}(b), we examine various networks from different domains for existence of diverse local patterns using $r_n^{local}$. In the two datasets, we witness skewed and multimodal distributions. We are essentially interested in how GNNs perform under different patterns, and further experimental analysis is provided in section 5.4.

\noindent\textbf{Observation 2.} \textit{The edge classification task relies on the different frequency signals between features of paired nodes.}

As discussed in Fig.~\ref{fig:view}, we believe that the implicit relational knowledge of the edge is related to the low and high-frequency information of the features of the paired nodes. In other words, the representation of the edge strongly depends on different frequency information between the connected nodes. GCN-based aggregation, as pointed out by many references \cite{Zhu2020,wu2019simplifying}, is essentially the smoothing of the features of connected nodes\cite{Nt19} to capture the similarities of graph structure and features. This causes implicit relationship patterns to become vague and indistinguishable in learned representation.

\section{Methods}

This section introduces the basic embedding units and self-supervised tasks to keep the original graph information.

\subsection{Pair and its neighbors}

To retain commonality and differentiation information between two connected nodes, we use pair as the basic modeling unit and take the feature information of paired nodes as a whole, avoiding smoothing its internal information. To further broaden the field of view of the model and use the correlation between the central unit and its domain information to obtain a more stable and robust effect, we introduce the "context" concept representing the aggregated neighboring feature patterns of a pair. 


\noindent\textbf{Ego features.} To represent $p_{u,v}$, we use the ego features from paired nodes $u$ and $v$. The $p_{u,v}$ is expressed with the direct concatenation of feature vector of nodes $u$ and $v$, ${X_{u,v}^{ego}=X_{u}^{ego}} \| X_{v}^{ego}$, $(X_{u}^{ego}=x_u, X_{v}^{ego}=x_v\in X)$. 

Here, "$\|$" represents the concatenation operation on the features of paired nodes $u$ and $v$ instead of "mixing" them with \textit{sum} or \textit{average} operation to avoid smoothing the high information between paired nodes. Thus, the relationships that highly rely on the feature patterns in the pair can be largely preserved, e.g., the relationships discussed in Fig.~\ref{fig:view}.

\noindent\textbf{Aggregated features.}
For the neighboring information, similar to the GNN-based solutions, we define an aggregation function to represent the surrounding context, denoted as $ X^{agg}_{u,v}$ of $p_{u,v}$. Here, the AGGR function aggregates node features only from the immediate neighbors of node $v$ or node $u$ (defined in the  $N(v), N(u)$) with permutation invariant functions, e.g., $mean$ or $sum$. As context might involve many different pairs, we use the $mean$ function to acquire low-frequency feature signals of a pair's surrounding context. Thus, the aggregated features represent the commodity around a pair. However, as the two paired nodes might have quite dissimilar contexts, the aggregation operations are taken separately towards each node and then concatenated to keep possible heterophily in between. Eq.~\ref{equ:agg} defines this process in aggregating features from a pair's context.

\begin{equation} 
\begin{aligned}
X^{agg}_{u,v} \gets Concat(&X^{agg}_u = AGGR(x_i,\forall i\in N(u)),\\&X^{agg}_v=AGGR(x_j,\forall j\in N(v)))
\end{aligned}
\label{equ:agg}
\end{equation}

The dimension of pair ego-features and aggregated-features is F, obviously, F = 2d.

\begin{figure}[h]
\centering
 \includegraphics[width=1.0\columnwidth]{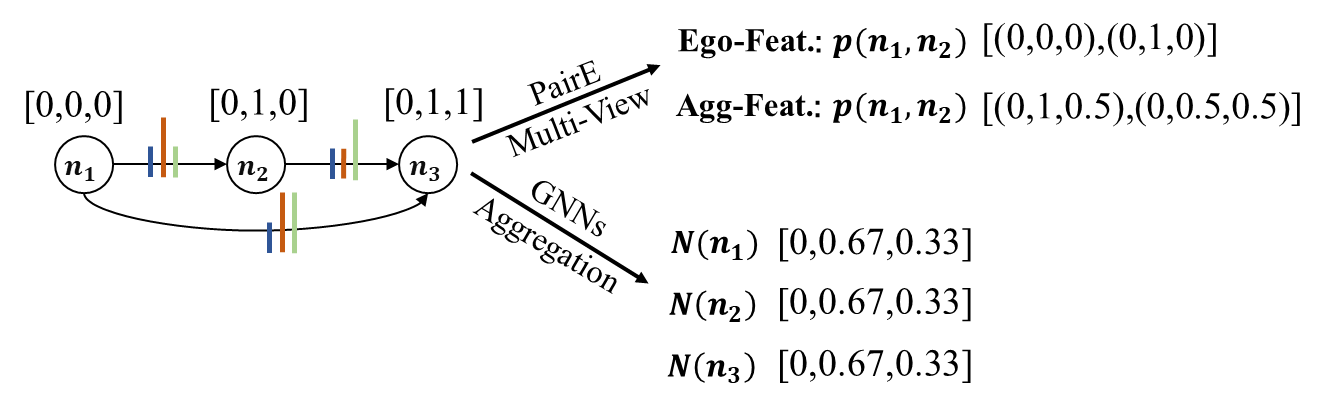}
\centering
\caption{PairE v.s. Existing GNNs: i) Different colors and line heights represent different dimensions of information and the strength of low-/high- frequency signals. ii) Explain the main difference between PairE and GNNs from the perspective of retaining the original low-/high- frequency signals.}
\vspace{-0.2cm}
\label{fig:compare}
\end{figure}

\noindent\textbf{Ego-/Agg-feat. vs Low-/High-freq. signals.} The differentiation between dimension-wise paired node features represents low-/high-frequency signals. We emphasize that different dimensions of node feature commonly can not be directly compared as they typically have diverse physical meanings, e.g., age and gender. As shown in  Fig.~\ref{fig:compare}, for $p_{n_1,n_2}$, the 2nd-dim features of $n_1, n_2$ with significant differences(0,1) can be seen as high-frequency, while the signal between the 1st/3rd-dim features are low-frequency. Thus, ego features contain both low and high-frequency signals. To reduce impacts from noise, we use the agg-features, mean aggregated features from $n_1,n_2$ neighbors to intensify the low-frequency signals and the mean-smoothed high-frequency signals between $n_1,n_2$ contexts. GNNs adopt neighbors aggregation iteratively and gradually smooth out high-frequency signals. It can be clearly seen from Fig.~\ref{fig:compare} that it will make the representations of all nodes close to or even completely consistent.

\subsection{Learning with different types of signals.} As GRL is to learn embeddings without label supports, and we need to find appropriate loss functions to realize such unsupervised learning. Generally, existing solutions can be divided into the reconstruction loss\cite{Veli17_GAT} or the contrast loss\cite{hamilton_grapshsage, Veli18_dgi}. As we have two different entities, we defined two different reconstruction tasks: a pair-based ego feature reconstruction task(denoted as ego-task) and a pair-based context reconstruction task(denoted as agg-task).
   
 \noindent\textbf{Ego feature reconstruction.} This task is designed to recover the pair's ego feature $X_{u,v}^{ego}$. Instead of reconstructing the actual values of ego features, we adopt the KL-divergence as our loss function by minimizing the Kullbach-Leibler distance $D_{KL}$ between the reconstructed feature distribution $Q({X}_{u,v}^{ego})$ and the prior distribution $P(X_{u,v}^{ego})$. By analogy with information theory, it is called the relative entropy of $P$ concerning  $Q$.  $ D_{KL}({P\parallel Q}) = 0 $ if and only if $P=Q$ almost everywhere. Here, our goal is to represent the distribution patterns rather than the actual feature values. 
 
 Here, our goal is to represent the distribution patterns rather than the actual feature values. In order to calculate the KL loss, we calculate the probability distribution of the $k$-th pair by Eq.~\ref{eq:softmax}. Where, $Q_k^{ego}(i)$ indicates the $i-{th}$ feature's distribution possibility in ego-task, and $H^{ego}$ represents the intermediate result of the output layer of ego-task. It is calculated by applying the standard exponential function to each element. $H^{ego}_k(i)$ and the normalized values are divided by the sum of all these exponents.

    \begin{equation} 
    \begin{aligned}
    \label{eq:softmax}
     Q_k^{ego}(i) =  \exp{\left(H^{ego}_k(i)\right)}  
      / & \sum_{j\ \in\ F}\exp{\left(H^{ego}_k(j)\right)} \\
      &  k\ \in\ E, i\ \in\ F
    \end{aligned}
    \end{equation}
    
    The loss function of KL is expressed by Eq.~\ref{eq:kl} :
    
\begin{equation}
\begin{aligned}
 \label{eq:kl}
\argmin & ({\rm D_{KL}}) = 
        &  Min\sum_{i\ \in\ E}\sum_{j\ \in F}{P_{i}\left(j\right)\ln\left(\frac{P_{i}(j)}{Q_i(j)}\right)}
\end{aligned}
\end{equation}


 \noindent\textbf{Aggregated feature reconstruction.}
 Similar to the pair-based ego feature reconstruction task, this task reconstructs the aggregated context features of the first-order neighbors $N(p_{u,v})$ of the pair $X_{u,v}^{agg}$, and uses a similar equation with Eq.~\ref{eq:softmax} for distribution calculation and a similar loss function based on KL-divergence to reconstruct $X_{u,v}^{agg}$. Due to the natural smoothness of the mean aggregation operation, $X_{u,v}^{agg}$ represents the information of the pair neighborhood features after smoothing. Therefore, this task provides contextual low-frequency information for the central pair, which can enhance the pair's ego features.

\begin{figure}[htb]
\centering
\begin{subfigure}{0.495\columnwidth}
\centering
\includegraphics[width=0.98\columnwidth]{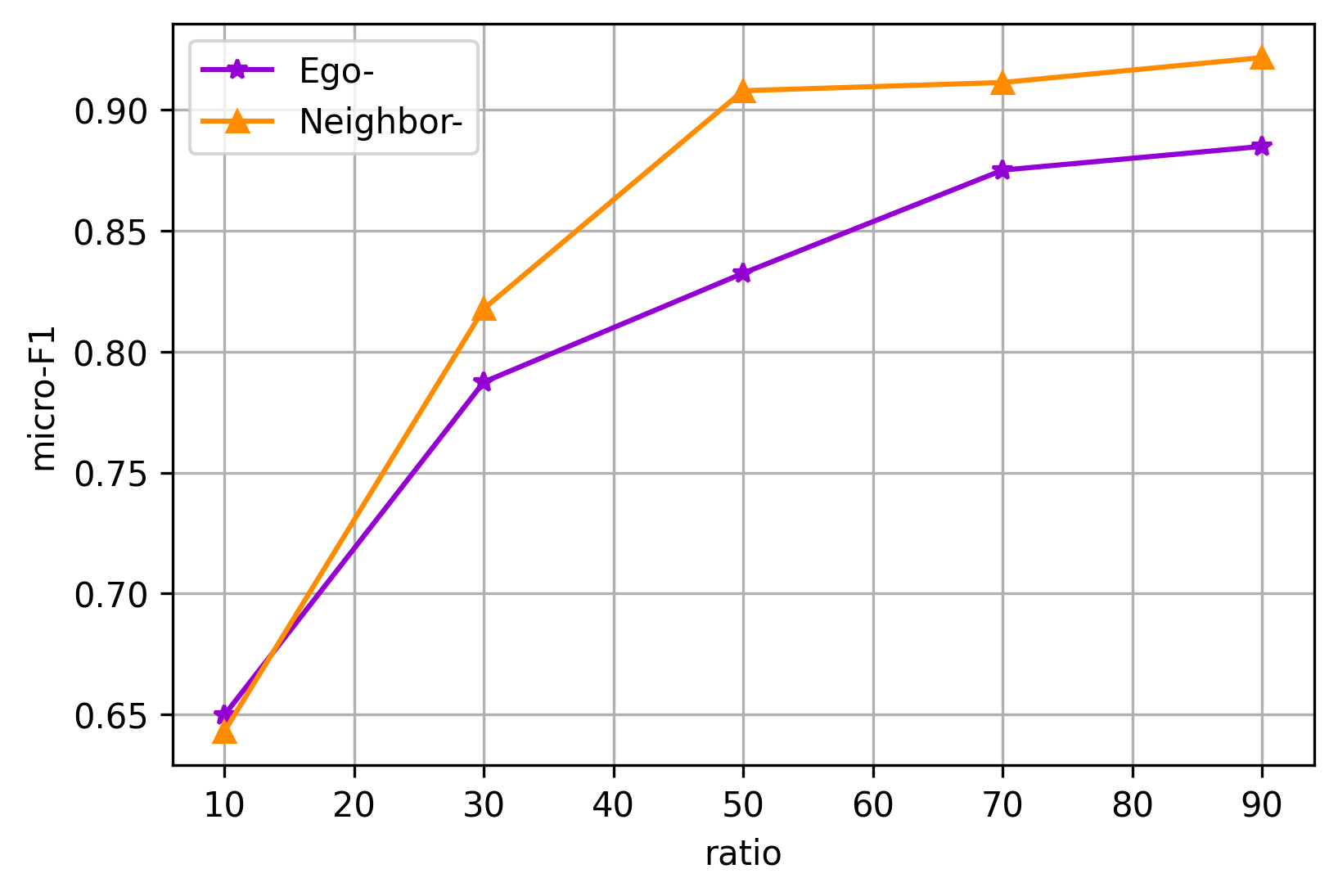}
 \caption{High Global Assortativity}
\end{subfigure}
\begin{subfigure}{0.495\columnwidth}
\centering
\includegraphics[width=0.98\columnwidth]{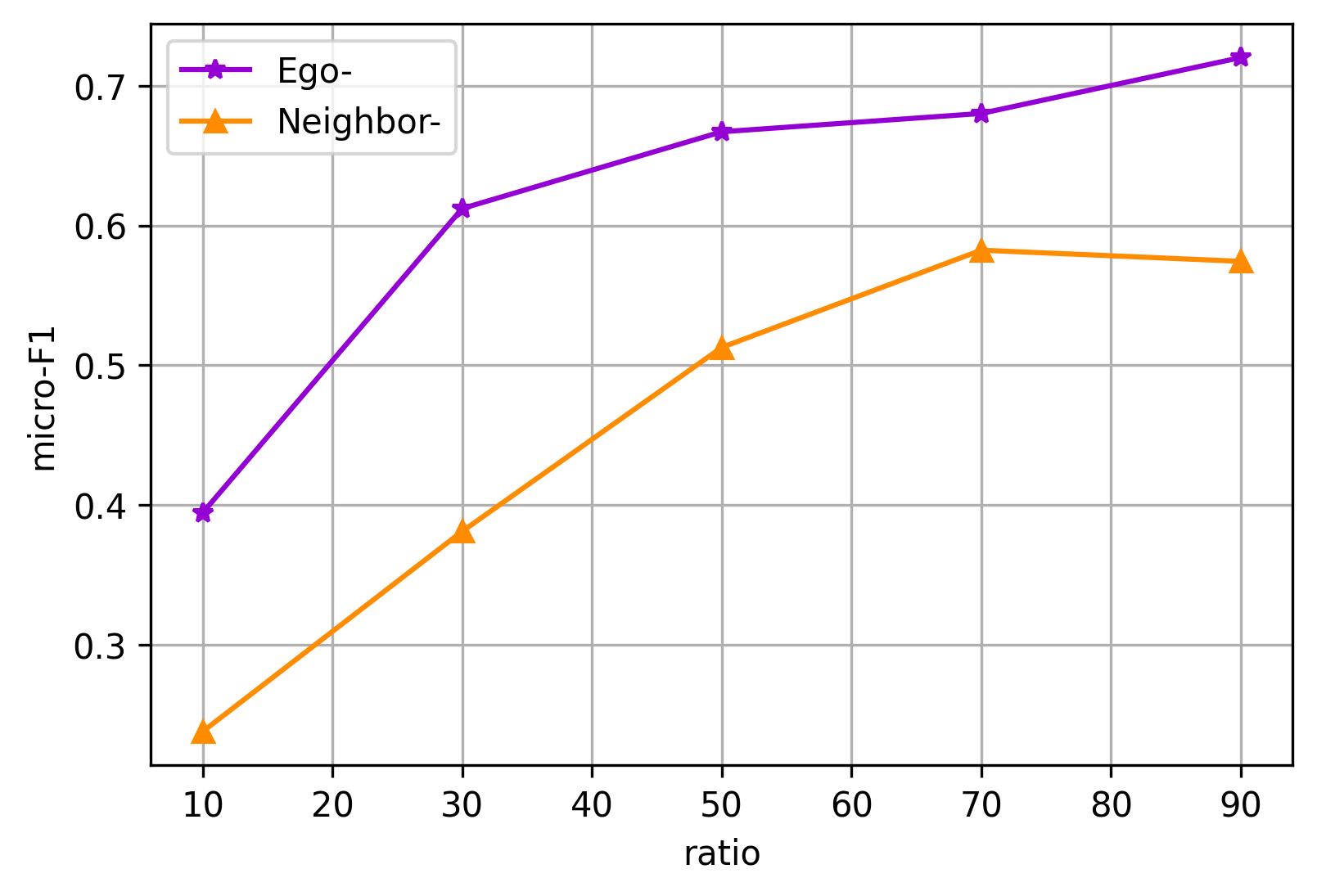}

 \caption{Low Global Assortativity}
\end{subfigure}
\caption{Performance comparisons for edge label classification with different global assortativity in FB15k-237}
\label{fig:ass_task}
\end{figure}

\begin{figure*}[ht]
\centering
\includegraphics[width=.95\textwidth]{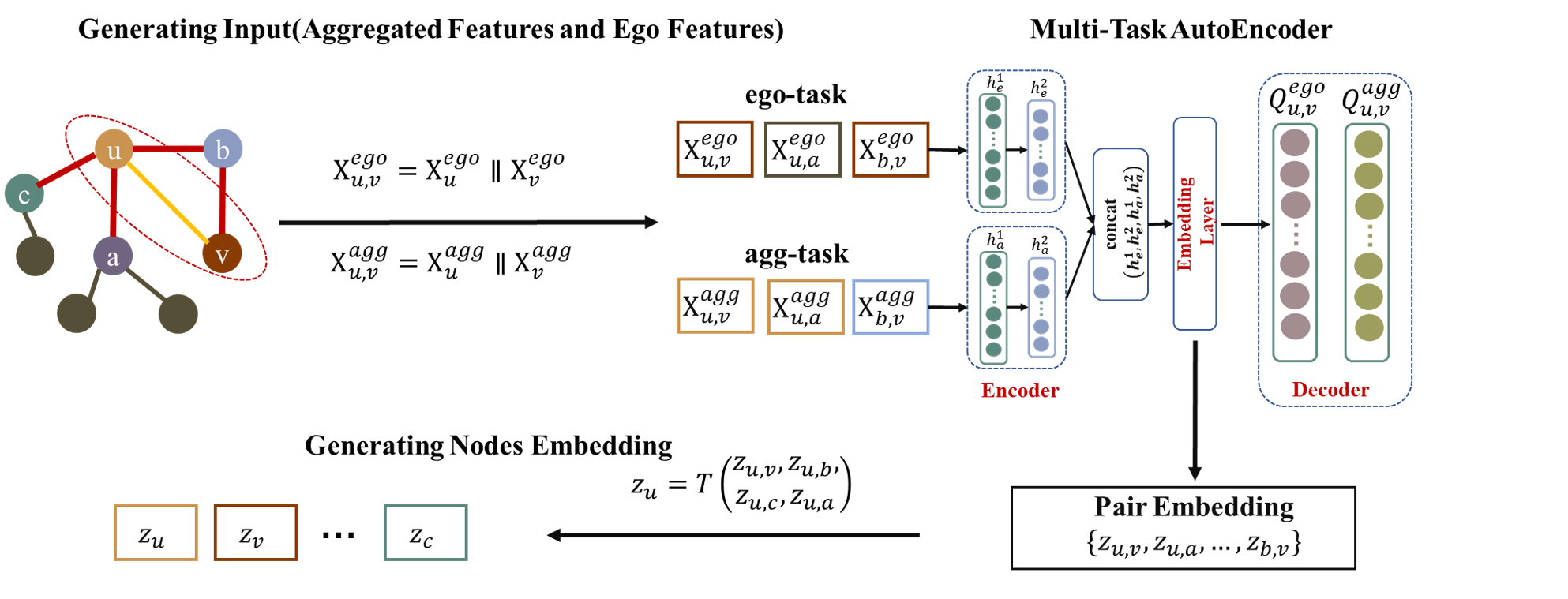}
\caption{A conceptual diagram for multi-self-supervised tasks autoencoder}
\label{fig:diagram}
\end{figure*}

\noindent\textbf{Network assortativity v.s. Reconstruction tasks.} Fig.\ref{fig:ass_task} shows the performance comparisons with embeddings calculated with the individual tasks with a simple AutoEncoder on edge classification with different label assortativity coefficients (edges with top/bottom three label coefficient). When the downstream tasks have high assortativity with similar edges in their neighborhood, the agg-task, by aggregating neighborhood information, essentially enhances the neighborhood's similarity (low frequency). Thus, it achieves better performance than using the pair ego features. 

When the downstream task exhibits disassortativity, most labels of neighboring edges are inconsistent with the pair's label. Thus, the feature distribution of the current pair normally is far away from its neighbors' aggregated ones. That distribution might contain a lot of information unrelated to the pair's label and can be seen as noises from this task's perspective. They can greatly disturb the learning process by making the representation deviate from the ground truth. In comparison, the ego-task guarantees the basic facts and can lead the agg-task with more than 25\% absolute improvement in Micro-F1. Fig. \ref{fig:ass_task}(a) and \ref{fig:ass_task}(b) clearly shows that for different downstream tasks, different types of information are needed. As GRL should be task-agnostic, it is important to integrate all those information into the embeddings.

\subsection{The designs of PairE}

According to the analysis in section 4.2, Ego feature reconstruction and Aggregated feature reconstruction capture the similarity (low-frequency) and difference (high-frequency) information from different perspectives in the graph, and the information of a single frequency cannot be adapted to have different assortativity level tasks. Therefore, to retain the information of the entire spectrum of low and high frequencies, we propose a novel multi-self-supervised tasks autoencoder to simultaneously learn a joint embedding space with data from multiple perspectives. Fig.~\ref{fig:diagram} describes the principal structure of the autoencoder. 
 
As shown in the right part of this figure, similar to other AutoEncoder solutions, it can be generally divided into three major parts, the \textit{Encoder}, \textit{Decoder} and the \textit{Embedding layer}. And PairE introduces three peculiar designs to support learning different frequency signals from respective tasks.

\noindent\textbf{Seperated En(de)code for respective tasks: } In this autoencoder, the two reconstruction tasks are encoded/decoded independently. Each is trained in two fully connected layers and only combined in the \textit{Embedding layer}. The two tasks are also respectively decoded by a fully connected layer with a softmax activation function defined in Eq.~\ref{eq:softmax}. Since different tasks focus on different frequency information in the graph, one shared embedding allows different signals to be integrated into one formal representation. Furthermore, this design exploits the mutual support from both tasks as a pair's ego features and its surrounding features are typically mutual-correlated. The impacts of decay or missing features, typical in many real-world networks, can be partially mitigated. Furthermore, learning with multiple tasks can generally improve performance compared to learning from a single task~\cite{Zhang17}. 

\noindent\textbf{Concatenation of intermediate representations: } Similar to the concatenation of the pair's ego features and agg features, we also separate the ego and context embeddings by encoding them separately. After aggregating the context representations, we combine the ego and the aggregated embeddings by concatenating rather than "mixing" all of them, similar to the GraphSage\cite{hamilton_grapshsage}. Choosing a concatenate function that separates the representations of pair-ego and pair-agg allows for more expressiveness\cite{Zhu2020}.
    
In addition to the concatenation, we combine the intermediate representations of ego and aggregated context at the final layer. Similar to the "add skip connections" operation introduced in ResNet\cite{He16}. It is later adopted in jumping knowledge networks for graph representation learning\cite{xu2018representation}. By combining all layers of the two tasks simultaneously, this design can preserve the integrity and personalization of the input information as much as possible. The inputs of the final layer are constructed as follows(As Fig. ~\ref{fig:diagram} shows):
 \begin{equation}
\begin{aligned}
  h_{u,v}^{emb} = Concat(h^{1}_e,h^{2}_e,h^{1}_{a}, h^{2}_{a})
\end{aligned}
\label{eq:concate}
\end{equation}

\noindent\textbf{Linear Activation Function:} Using  graph-agnostic dense layers to reconstruct the original features of each pair: $h_{u,v}^{self} = \sigma(X_{u,v}^{self} W)$, where $\sigma$ is an activation function, $W$ is a weight matrix. Here, the linear function is used in this paper. By performing an only linear transformation on the input features, our solution will not smooth out high-frequency signals in the ego features of the pair. 


\noindent\textbf{Discussion:}  
We analyze the expressive power of PairE and GCN-based approaches from the perspective of the similarity between edge embeddings. For any two edges, $(u,v),(u,v^{'})\in E$, with a common node $u$. $P(X),P(R),Q(X)$ are the orignal feature distribution, embedding distribution, and the reconstructed distribution. In order to evalute the similarity between vectors, the Cosine similarity measure is used to measure of similarity between two non-zero vectors of an inner product space. Let $\mathcal{S},\mathcal{S}_{P}$ denote the Cosine similarity of edge features, learned edge embeddings by PairE. Values range between -1 and 1, where -1 is diametrically opposed, 0 means independent, and 1 is exactly similar.

$$\mathcal{S}=\frac{P(X_{u,v})\cdot P(X_{u,v^{'}})}{|P(X_{u,v})||P(X_{u,v^{'}})|}, \mathcal{S}\in[-1,1]$$



\noindent\textbf{Proposition 1.} The embeddings learned by PairE can be as close as possible to the original distribution of the feature. 



\noindent\textit{Proof.} The goal of the feature reconstruction make the final reconstructed feature distributions of pair u,v $Q(X_{u,v})$ is nearly identical to the original feature distribution $P(X_{u,v})$.
   $$ argmin(D_{KL}({P\parallel Q})) \rightarrow 0 \Rightarrow Q(X_{u,v}) \rightarrow P(X_{u,v}) $$
  
The $Q(X_{u,v})$ are reconstructed with the generated edge embeddings $R_{u,v}$, thus, $R_{u,v}$ have the same distribution as $Q(X_{u,v})$. Thus, we can deduce: $P(R_{u,v}) \rightarrow P(X_{u,v})$. It means that the generated representation $R_{u,v}$ will have the similar distribution as the original paired feature distribution if the KL divergence approaching zero. Therefore, embeddings retain a relatively complete relationship pattern in the original feature. 

Therefore, it is easy to calculate: 
   $$\mathcal{S}_P = \frac{P(R_{u,v})\cdot P(R_{u,v^{'}})}{|P(R_{u,v})||P(R_{u,v^{'}})|} \rightarrow \frac{P(X_{u,v})\cdot P(X_{u,v^{'}})}{|P(X_{u,v})||P(X_{u,v^{'}})|} = \mathcal{S}$$
Where we can clearly see that $\mathcal{S}_P$ will gradually approach $\mathcal{S}$ during the learning process of PairE, in other words, PairE can be seen as a full-pass filter on the input signals with certain compression.

\noindent\textbf{Lemma 1.} Most existing GNNs, e.g., GCN, based a low-frequency filter: $(D+I)^{-1/2}(A+I)(D+I)^{-1/2}$, and only have the capability to narrow the distance between node representations, where $D,I,A$ denote the diagonal degree matrix, the identity matrix, and the adjacency matrix of $G$, respectively:  
$$\frac{P(R_{u})\cdot P(R_{v})}{|P(R_{u})||P(R_{v})|} \textgreater \frac{P(X_u)\cdot P(X_v)}{|P(X_u)||P(X_v)|} $$

As this fact has been discussed in many papers, we omitted the detailed proof. Detailed proof can be found in reference~\cite{bo21}.

For the edge-related tasks, most GNNs need to obtain edge representations based on node representations with certain conversion operations, e.g, sum, mean, concatenate. For fair comparison, we adopt concatenate operation as an example: $P(R_{u,v}) = P(R_{u})||P(R_v)$. Obviously, edge representations still obey Lemma1.
$$\mathcal{S}_G = \frac{P(R_{u,v})\cdot P(R_{u,v^{'}})}{|P(R_{u,v})||P(R_{u,v^{'}})|} \textgreater \frac{P(X_{u,v})\cdot P(X_{u,v^{'}})}{|P(X_{u,v})||P(X_{u,v^{'}})|}$$



In addition, as the number of iterative training increases, the discrimination between representations is gradually reduced. Thus, it is easy to cause the so-called over-smoothing problem: $\mathcal{S}_G \rightarrow 1$. Due to a lack of distinction in the reconstrued edge representations, it isn't easy to make correct inferences about the relationship between nodes.

\subsection{Edge to node embeddings}
To support node-related downstream tasks, we need to define a translator function $T$ to translate pair embedding to node embeddings. This function should be permutation-free and several operations can be used: \textit{min}, \textit{max},\textit{ mean}, and \textit{sum}. As pointed out in \cite{Epasto19}, one node is a composite entity with multiple roles. Thus, the \textit{sum} translator defined in Eq.\ref{eq:sum}. With this translator, the embedding of node $u$, defined as $z_u$, is the sum of the set of pair embedding $z_{u,v_i}$ with $u$ as a starting point. The appendix provides performance comparisons with different types of translators.


\begin{equation} 
  z_u = Sum\left(z_{u,v_i},v_i \in N(u)\right)
 \label{eq:sum}
\end{equation}

\subsection{Time and space complexity}
According to Fig.~\ref{fig:diagram}, the complexity of PairE mainly depends on the number of pairs $|E|$ and the dimension of pair features $F$. For the computation complexity, every pair is calculated O(epoch) times, then the total computational complexity here is $O(epoch\cdot|E|\cdot F)$.
PairE only needs to put the concatenated features as inputs, and the space occupation is much smaller than adjacent vectors.  The space consumption for training is related to the batch size $b$, and the dimension of pair features $F$, about $O(2\cdot b\cdot F)$. As PairE is mini-batch training and we do not need to store intermediate embeddings in the GPU. Thus, PairE can train massive graphs with GPU with small memory.

\section{Experiments}
\label{sec:experiment}

The performances of PairE are tested on both edge-related tasks and node-related tasks to answer three questions: 

\noindent{\textbf{Q1:} Can PairE effectively encode different types of signals into the embeddings from the perspective of different entities to support various downstream tasks?}

\noindent {\textbf{Q2:} What role has the ego and agg tasks played in keeping different types of signals? }

\noindent{\textbf{Q3:} Whether PairE is computation efficient and resources scalable to be used in large graphs?}

\subsection{Datasets}
\label{sec:setup}

To investigate the learning ability of PairE towards the complex structures from graphs in different application domains, a variety of real-world graph datasets from benchmarks are selected. The standard benchmark dataset:
\begin{itemize}

  \item \textit{Cora, Citeseer, Pubmed}: standard citation networks where nodes represent documents, edges are citation links, and features are the bag-of-words representation of the document. 
  \item \textit{Cornell, Wisconsin}: two webpage datasets, where nodes represent web pages, and edges are hyperlinks between them. Node features are the bag-of-words representation of web pages. The web pages are manually classified into five categories, student, project, course, staff, and faculty. 
  \item \textit{PPI:} multi-layer datasets of PPI from biology for the interactions between different proteins. \item \textit{DBLP}: from collaboration network. In PPI and DBLP, each node might play multiple roles and have multiple labels\footnote{https://linqs.soe.ucsc.edu/data}. 
  \end{itemize}
For the networks with edge labels, four datasets are selected: 
  \begin{itemize}
  \item \textit{Cuneiform}: the Cuneiform for handwriting Hittite cuneiform signs; 
 \item \textit{FFtwyt}\footnote{http://multilayer.it.uu.se/}: a friend feed of Twitter and Youtube in the social network;
 \item \textit{FB15K-237} and \textit{WN18RR} \footnote{https://github.com/villmow/datasets\_knowledge\_embedding}: two wildly used knowledge graphs with  multi-labe edges.
 \end{itemize}

In those four networks, each edge might have one or multiple types. Since many knowledge graphs have no node features, the node embeddings generated by DeepWalk are used instead. Tab.~\ref{tab:datasets} shows their basic properties. Please note we perform unsupervised learning. Labels of all nodes and edges are hidden during learning.

\begin{table}[htbp]
 \centering
  \caption{Datasets Properties, ME denotes Multi-label Edges and MN denotes Multi-label Nodes}
    \begin{tabular}{lrrrrr}
    \toprule
    \textbf{Dataset}  & \textbf{Nodes} &  \textbf{Edges} & \textbf{Node T.} & \textbf{Edge T.} & \textbf{Feat.} \\
    \midrule
    Cora &2708 & 5429& 7& -&1433 \\
    CiteSeer &3327 & 4732& 6& -&3703\\
    Pubmed  & 19,717  & 44,327  &  3  & - & 500\\
    Cornell &183 &280 & 5& -& 1703\\
    Wisconsin & 251&466 & 5& -& 1703\\
    DBLP  &  28,702 & 68,335 & MN(4) & - & 300\\
    PPI  &  56,944   &  818,716  &  MN(121)  &  - & 50 \\
    Cuneiform  & 5,680  & 23,922  &  MN(7)  & 2 &  3 \\
    FFtwyt  & 6,407  & 65,235   &  - & ME(3) &  128(dp) \\ 
    WN18RR  & 40,943  & 92,879  & -  & ME(11)  & 128(dp) \\
    FB15K-237  & 14,541  & 283,868  & -  & ME(237) & 128(dp) \\
    \bottomrule
    \end{tabular}
  \label{tab:datasets}
\end{table}%

\definecolor{mygray}{gray}{0.85}

  \begin{table*}[h]
  \begin{center}
  \caption{Edge classification results, trained with 30\% ratio, evaluated with average Micro-F1. Analogous trends hold for averaged Macro-F1 scores. Datasets with + are of multi-label edges. \colorbox{mygray}{Gray background} and \textbf{bold font} represent the best results in all algorithms and unsupervised settings, respectively. "--" denotes F1-Score is less than 0.1}
  \label{tab:edge}
  \renewcommand\arraystretch{1.1}
    \begin{tabular}{p{4.5em}|c|c|c|c}
    \hline
    Name&Cuneiform&FFtwy$t^+$ &FB15K\-$237^+$ &WN18RR$^+$ \\
    $r_e^{global}$ &$0.26$ &$0.07\sim0.53 $ &$0.001\sim0.86$ &$0.13\sim0.92$  \\
    \hline
    DeepWalk &84.24$\pm$ 2.05&65.59$\pm$ 0.52 & 62.30$\pm$ 0.49 & 26.56$\pm$ 2.00\\
    ARVGE& 72.52$\pm$0.62 & 76.17$\pm$0.96 & 56.92$\pm$0.98 & 64.80$\pm$0.63\\
    TansR&70.29$\pm$0.65 &80.19$\pm$0.42 &78.48$\pm$0.43 &71.65 $\pm$ 0.87\\
    SAGE-U &71.78$\pm$0.42 &77.54$\pm$1.13 &56.14$\pm$0.78 & 46.19$\pm$2.91\\
    DGI&71.08$\pm$0.11 &69.84$\pm$5.76 & 35.04$\pm$14.10 & 36.36 $\pm$9.11 \\
    NodeE&73.34$\pm$2.37 &70.30$\pm$0.65 & 42.64$\pm$0.70 &54.88 $\pm$ 0.42\\
    PairE&\textbf{95.71$\pm$1.54} &\colorbox{mygray}{\textbf{83.20$\pm$ 0.52}} &\colorbox{mygray}{\textbf{89.36$\pm$0.62}} & \colorbox{mygray}{\textbf{74.99 $\pm$ 0.53}} \\
    \hline
    GCN&\colorbox{mygray}{98.52$\pm$ 0.62} & 78.71$\pm$ 0.23 & -- & 46.40$\pm$1.17 \\
    SAGE-S & 92.17$\pm$ 1.06 & 81.45$\pm$0.32 & -- & 42.48$\pm$1.74 \\
    PGE & 71.35$\pm$0.66 & 79.56 $\pm$ 0.21 & -- & 54.84$\pm$0.83 \\
    FAGCN& 95.01$\pm$4.90	&77.74$\pm$1.39	& -- & 35.21$\pm$6.09 \\
    H2GCN& 80.19$\pm$0.27 & $72.81\pm0.09$& -- & 29.66$\pm$0.95\\
    \hline
    \end{tabular}
    \end{center}
\end{table*}

  \begin{table*}[!htb]
  \begin{center}
  \caption{Node classification results, trained with 30\%, evaluated with average Micro-F1. Datasets with * are of multi-label nodes. Different marks have the same meaning as Tab~\ref{tab:edge}}
  \label{tab:node}
  \renewcommand\arraystretch{1.1}
    \begin{tabular}{p{4.5em}|c|c|c|c|c|c|c|c}
    \hline
    Name&Cora&Citeseer&Pubmed&Cornell&Wisconsin &DBLP*&PPI*&Cuneiform* \\
    $r_n^{global}$ &$0.77$ &$0.67$ &$0.69$ &$0.07$ &$0.05$ &$0.58\sim0.78$ &$0.06\sim0.17$ &$0.07\sim0.78$ \\
    \hline
    DeepWalk &80.05$\pm$  5.74&56.02$\pm$10.14 & 79.80$\pm$3.66 & 48.67$\pm$27.84 & 48.67$\pm$27.84 & 70.00$\pm$2.52 & 51.56$\pm$0.58 &49.11$\pm$1.40 \\
    ARVGE& 79.06$\pm$7.94 & 60.42$\pm$10.01 & 71.50$\pm$3.18 & 54.33$\pm$30.27 & 54.42$\pm$21.84 & 77.74$\pm$0.84 & 65.79$\pm$1.37 & 54.73$\pm$0.83 \\
    TansR&60.3$\pm$12.70 &40.58±8.22 &63.48$\pm$1.76  & $42.00\pm16.18$ &$49.29\pm28.31$ & 59.87$\pm$0.37  & 52.28$\pm$0.79 & 50.85$\pm$1.31 \\
    SAGE-U&80.18$\pm$9.85 &72.96$\pm$10.59 &83.82$\pm$3.01 & 48.18$\pm$13.61 & 60.00$\pm$8.43 & 79.52$\pm$0.76 & 59.21$\pm$0.99 & \underline{70.59$\pm$1.02} \\
    DGI&81.28$\pm$7.58 &68.94$\pm$9.1 & 83.11$\pm$1.50 & 38.73±11.93 & 60.00$\pm$22.31 & 78.34$\pm$0.68 & 89.58$\pm$0.58 & 50.18$\pm$3.02 \\
    
    CAN&82.38$\pm$6.99 &66.73$\pm$8.23 &79.24$\pm$2.23 &42.36$\pm$20.43 & 46.67$\pm$8.43 & 24.43$\pm$2.67 & 49.91 $\pm$0.46 & --\\
    NodeE&79.43$\pm$5.79 &69.54$\pm$7.89 & 87.96$\pm$2.15 & 57.67$\pm$10.77 & 66.43$\pm$23.46 & \underline{80.05$\pm$1.02} & \underline{91.48$\pm$0.25} & 43.13$\pm$1.24 \\
    PairE&\textbf{86.51$\pm$8.52} &\textbf{75.53$\pm$5.58} &\textbf{88.57$\pm$2.72} & \textbf{66.73$\pm$7.77} & \textbf{68.00}$\pm$\textbf{9.98} & \colorbox{mygray}{\textbf{80.58}$\pm$\textbf{0.62}} &\colorbox{mygray}{\textbf{94.83}$\pm$\textbf{0.18}} & \colorbox{mygray}{\textbf{75.12}$\pm$\textbf{0.78}} \\
    \hline
    GCN& 87.33$\pm$2.05 & \colorbox{mygray}{78.37$\pm$4.16} & 87.37$\pm$0.64 &$65.99\pm10.19$ & $58.33\pm5.27$ & 79.53$\pm$0.45 & 41.86$\pm$0.39 & 47.42$\pm$0.85 \\
    SAGE-S& \colorbox{mygray}{87.92$\pm$3.00} & 76.02$\pm$2.08	& \colorbox{mygray}{89.27$\pm$0.75}	& \colorbox{mygray}{$76.0\pm8.0$} & $78.33\pm10.0$ & 79.86$\pm$0.40 & 40.44$\pm$0.64	& 51.02$\pm$0.69\\
    
    PGE&60.27$\pm$2.64 & 52.97 $\pm$3.43 & 85.03 $\pm$0.39 & 54.35$\pm$9.99 & 52.63 $\pm$12.48 &78.23$\pm$0.35 & 63.97$\pm$0.31 & 62.52$\pm$1.73 \\
    FAGCN& 85.96$\pm$1.51	&74.14$\pm$1.91	& 86.83$\pm$0.44 & $72.59\pm8.63$& $77.36\pm6.35$ & 67.91$\pm$1.69	&39.24$\pm$0.97	&25.68$\pm$1.79\\
    H2GCN& 85.69$\pm$1.76 & 74.95$\pm$2.38 & 87.28$\pm$0.52 & $71.85\pm6.45$ & \colorbox{mygray}{$78.94\pm5.52$} & 75.99$\pm$0.40 &47.65$\pm$0.25 &24.26$\pm$2.14\\
    \hline
    \end{tabular}
    \end{center}
\end{table*}
  
\subsection{Baselines}

PairE is an unsupervised network embedding algorithm. To make a fair comparison,  we have selected only the following state-of-the-art node embedding algorithms, which are unsupervised in nature and have publicly available source code. They are classified into two categories: methods based on topology and node features. Furthermore, to best demonstrate the effectiveness in different assortativity of networks, we also compare four semi-supervised solutions optimized for different types of networks.

\noindent  \textbf{Topology only}:
\begin{itemize}
\item  DeepWalk(denoted as DW) ~\cite{Perozzi14}, the first approach for learning latent representations of vertices in a network with NLP. 
\item  {TransR}~\cite{TransR15} builds entity and relation embeddings in separate entity space and relation spaces to support N-N relations. 
\end{itemize}

\noindent  \textbf{Both topology and node features}:
\begin{itemize}
  \item {ARVGE} ~\cite{ARGA18} is an autoencoder-based solution that enhances GAE to seek robust embedding by adding an adversarial module on the obtained embeddings. This method is known to have cutting-edge performance in the link prediction task.
  \item {SAGE-U} ~\cite{hamilton_grapshsage}: an unsupervised GraphSage variant that leverages node feature information to generate node embeddings by sampling a fixed number of neighbors for each node. 
  \item {DGI}~\cite{Veli18_dgi}: a contrastive approach via maximizing mutual information between patch representations and high-level summaries of graphs.
  \item {CAN}~\cite{meng2019co}: a variational auto-encoder that embeds each node and attribute with means and variances of Gaussian distributions.
  \item  NodeE is used for ablation studies, and it works on the node level with the same autoencoder as PairE.

\end{itemize}
\noindent \textbf{Semi-Supervised Algotithms}:
\begin{itemize}
\item{GCN}~\cite{Kipf16}: essentially uses a low-pass filter to retain the commonality of node features and fails to generalize to networks with heterophily (or low/medium level of homophily).
\item{SAGE-S} ~\cite{hamilton_grapshsage}: a semi-supervised variant of GraphSage that uses the same low-pass filter as GCN.
\item{PGE}~\cite{hou2019representation}: a graph representation learning framework that incorporates both node and edge
properties into the graph embedding procedure.
\item{H2GCN}~\cite{Zhu2020}: identifies a set of key designs that can boost learning from the graph structure in heterophily without trading off accuracy in homophily.
\item{FAGCN}~\cite{Zhu21}: frequency adaptation GCN with a self-gating mechanism, which can adaptively integrate both low and high-frequency signals in the process of message passing.
\end{itemize}

The semi-supervised solutions are trained in an end-to-end fashion and optimized for specific downstream tasks with the support of downstream labels. Thus, it is typical that semi-supervised solutions might have better performance than unsupervised ones.  

Here, only limited baselines on several representative graphs are reported. More comprehensive experiment results with more baselines, e.g., TADW~\cite{yang2015TADM}, ProNE~\cite{Zhang19_prone}, and datasets, e.g., Proteins\_full are provided in the Appendix. 
 
\subsection{Experiment settings}

In this section, experiment setups for different downstream tasks and baselines are defined. 

\noindent\textbf{Node/Edge classification task:} For these tasks, all the nodes are used in the embedding generation process. For PairE, the node embeddings are generated with Eq. ~\ref{eq:sum}. Edge embeddings for all compared baselines are calculated with the $L2$ norm formula of two-node embeddings. The embeddings of vertices/edges from different solutions are taken to train a classifier with different training ratios from 30\%, 50\% to 70\%, and classification accuracy is evaluated with the remaining data. Due to space limitations, Tab.~\ref{tab:edge} and Tab.~\ref{tab:node} only show the results with the 30\% train ratio for both unsupervised and semi-supervised solutions. For each network, we used 10 random splits, and the average performances in Micro-F1 are reported. 

\noindent\textbf{Link-prediction task:} We use a similar prepossessing step in~\cite{Grover16}. Our experiments held out a set of 15\% edges for testing and trained all models on the remaining subgraph. Additionally, the testing set also contains an equal number of non-edges. In this task, the edge embeddings for all methods are calculated from the source and target node embeddings with the $L2$ operator. 

\noindent\textbf{Experiment settings:} For all semi-supervised algorithms and partially unsupervised algorithms(Sage-U, DGI, and ARVGE), we use the most commonly used hyperparameter settings: Implementing in Pytorch with Adam optimizer, we run 10 times and report the mean values with standard deviation. The hidden unit is fixed at 16 in all networks. The hyper-parameter search space is: learning rate in $\{0.01,0.005\}$, dropout in $\{0.4, 0.5, 0.6\}$, weight decay in {1E-3, 5E-4, 5E-5}, number of layers in $\{1, 2,..., 8\}$, scaling hyper-parameter in $\{0.1,...,1.0\}$. Besides, we run 200 epochs and choose the model with the highest validation accuracy for testing.


The embedding dimensions of both pair embeddings in PairE and node embeddings for other unsupervised baselines are set to 128. Other parameters are set as the default settings in the corresponding papers. For the PairE, we set epoch 30 and batch size 1024. A classifier with the logistic regression algorithm is used with the one-vs-rest strategy. We report results over 10 runs with different random seeds and train/test splits, and the averaged ROC\_AUC/Micro-F1 score is reported. The model configurations are fixed across all the experiments. Detailed settings are omitted. 

\subsection{Performance comparisions}
\label{sec:results}
   
\noindent\textbf{Edge classification Task:} This task aims to answer Q1. As shown in Tab.~\ref{tab:edge}, PairE outperforms by a significant margin in all four tested datasets across all tested ranges. For unsupervised algorithms, PairE achieves more than 13.6\%$\sim$ 36\% performance edges in Cuneiform and 43$\sim$155\% in FB15K-237 and 5$\sim$182\% WN18RR. On average, PairE achieves up to 101.1\% improvement on this task. Although the node features are from deepwalk embeddings, PairE can perform better than all the other baselines, including DW.
In comparison, TransR ranks second in Micro-F1 as it is designed to build the relations between nodes with a transformation function. ARVGE is optimized for the link-prediction task focusing on link existence estimation, while this task demands more expression power as edges are with multiple labels. In our experiences, other Trans and ARVGE variants also display similar performance.

It is particularly worth noting that the classification effect of PairE on multi-label datasets far surpasses the comparison of semi-supervised benchmarks(even for H2GCN and FAGCN) optimized for homophily and heterophily) and achieves the best performance. 
These results clearly show an interesting fact that the edge has strong correlations with both the node features and the context of the paired nodes. PairE can differentiate those structural differences at a finer granularity and identify heterogeneous information, leading to high-quality pair representations with superior performance. 

Meanwhile, an interesting phenomenon is that all those semi-supervised methods fail on the FB15k-237 dataset. We guess that each edge in the dataset has a 237-dimensional label and the assortativity of most labels is extremely low. The relationships among features and labels are too complicated to be effectively learned by those semi-supervised solutions. 

\noindent\textbf{Node classification Task:} The task provides answers to Q2. The node embeddings of PairE are translated with Eq.~\ref{eq:sum}. As shown in Tab.~\ref{tab:node}, under unsupervised settings, PairE still outperforms those baselines by a significant margin in all eight tested datasets with different node label assortativity compared with other unsupervised strong baselines. And on average, up to 82.5\% performance gain has been achieved. Compared to the semi-supervised baselines, PairE achieves comparable performance even with no node labels during embedding. 
For the datasets with high node label assortativity, the semi-supervised solution GCN and Sage-S achieves good performance as GCN and Sage-S are optimized for homophily networks, e.g., Cora. The FAGCN and H2GCN are generally designed for heterophily networks, and they reach the best performance in Cornell and Wisconsin, whose nodes assortativity are only 0.07 and 0.05, respectively. Sage-S have a strong performance in both kinds of networks as it updates the node ego-embeddings by concatenating the aggregated neighbor-embeddings.

In the graphs with multiple types of node labels, e.g., DBLP, PPI, and Cuneiform, PairE displays significant performance advantages over the other ten baselines, including both unsupervised and semi-supervised solutions. In PPI, PairE achieves the averaged Micro-F1 score to 0.94 and up to 141.7\% relative performance improvement. PairE has a similar performance advantage in Cuneiform as both datasets have very low node label assortativity. As pointed out in \cite{Epasto19}, one node might play multiple roles, which the pair embeddings can partially support. Those complex correlations can be used to support the multi-label node classification task better. However, even for semi-supervised solutions, existing solutions fail to learn a model that can effectively represent the labels with highly different assortativity values. One exception is DBLP. Its labels are highly correlated and have similar high assortativity values(0.58$\sim$0.78); thus, different baselines display comparably good performance in this dataset. 

Combined with Tab.~\ref{tab:edge}, we can see that PairE has very consistent performance in both the node and the edge classification task. It is important for an unsupervised embedding solution if its embedding can be used for many different and possible unknown purposes. PairE obviously has such excellent characteristics. Furthermore, results also show that the semi-supervised GNN solutions face significant limitations in learning at datasets with labels of diverse assortativity in both node and edge classification tasks.

\begin{figure}[tb]
\centering
 \includegraphics[width=3.2in,height=2.in]{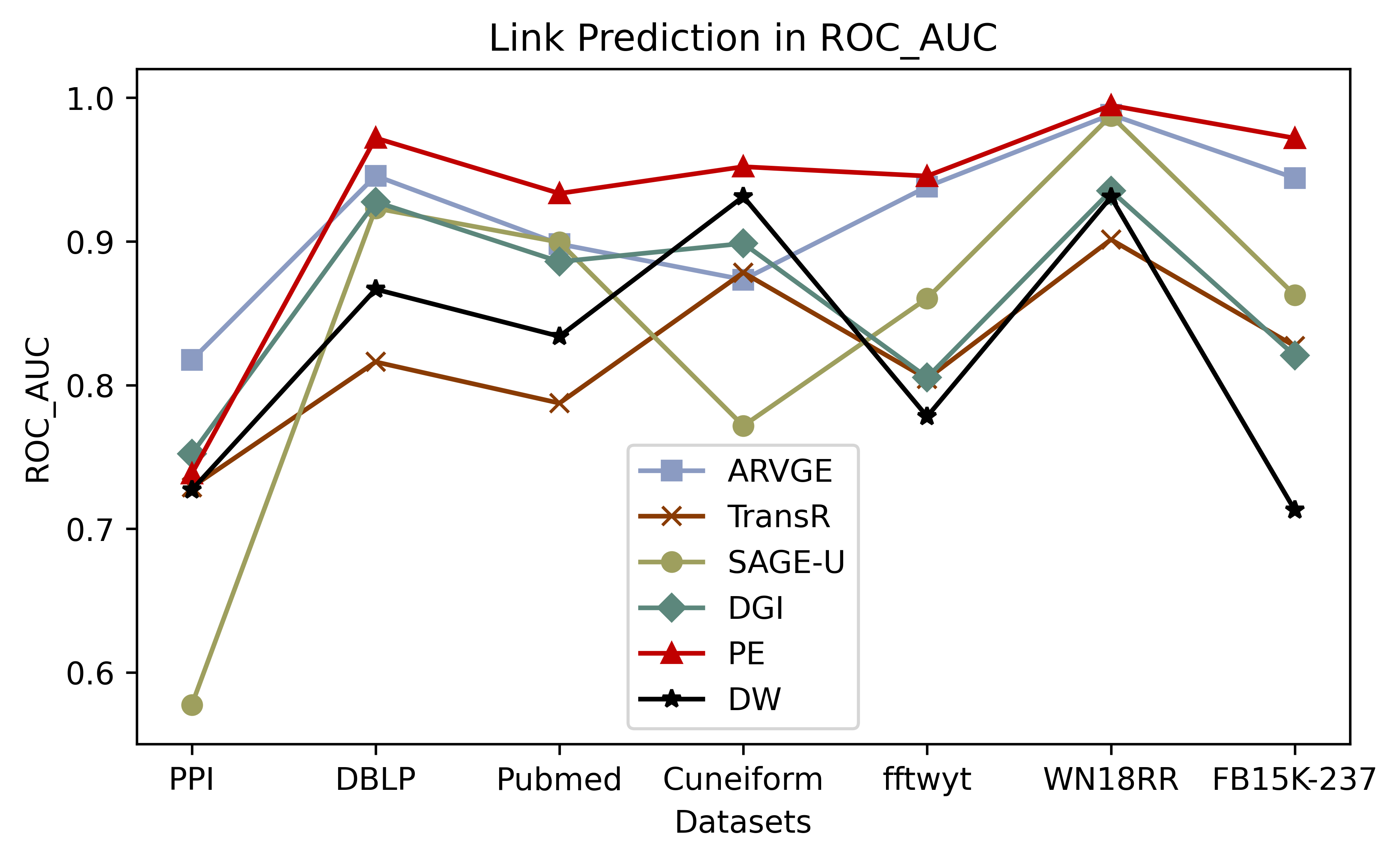}
  \vspace{-0.3cm} 
\centering
\caption{Comparisons in the Link Prediction tasks, evaluated in ROC\_AUC }
\vspace{-0.3cm}
\label{fig:link}
\end{figure}

\noindent\textbf{Link Prediction Task:} Fig.~\ref{fig:link} shows the results for the link prediction task in ROC\_AUC. Since ARVGE reconstructs graph structure by directly predicting whether there is a link between two nodes, it has outstanding performance in this task. Nevertheless, PairE still beats these very competitive benchmarks in 6 out of 7 datasets in ROC\_AUC. We observe that two nodes matching neighboring patterns, e.g., close together or matching features, are more likely to form a link. TransR, due to its lack of usage of node features, suffers the most in this task. Results in AP also display similar trends. 
  
Here, the only exception is PPI. It is a protein interaction network, and different amino acid types are more likely to connect. Thus, there are rich high-frequency signals between the features of paired nodes. As the link prediction task demands fundamental relations between nodes, the rich information might pose complexities to the classifier. The following ablation studies provide further supports for this phenomenon.

\noindent \textbf{Node Classification vs. Local Assortativity:} In order to better show the difference of nodes with different local assortativity,  an analytical experiment is introduced for Pubmed(homophily) and Wisconsin(heterophily).  The setting of the rask is consistent with the node classification task.

\begin{figure}[htb]
\centering

\begin{subfigure}{0.8\columnwidth}
\centering
\includegraphics[width=0.98\columnwidth]{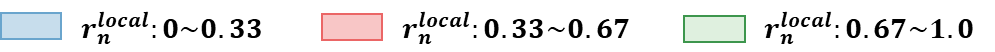}
\end{subfigure}

\begin{subfigure}{0.495\columnwidth}
\centering
\includegraphics[width=0.98\columnwidth]{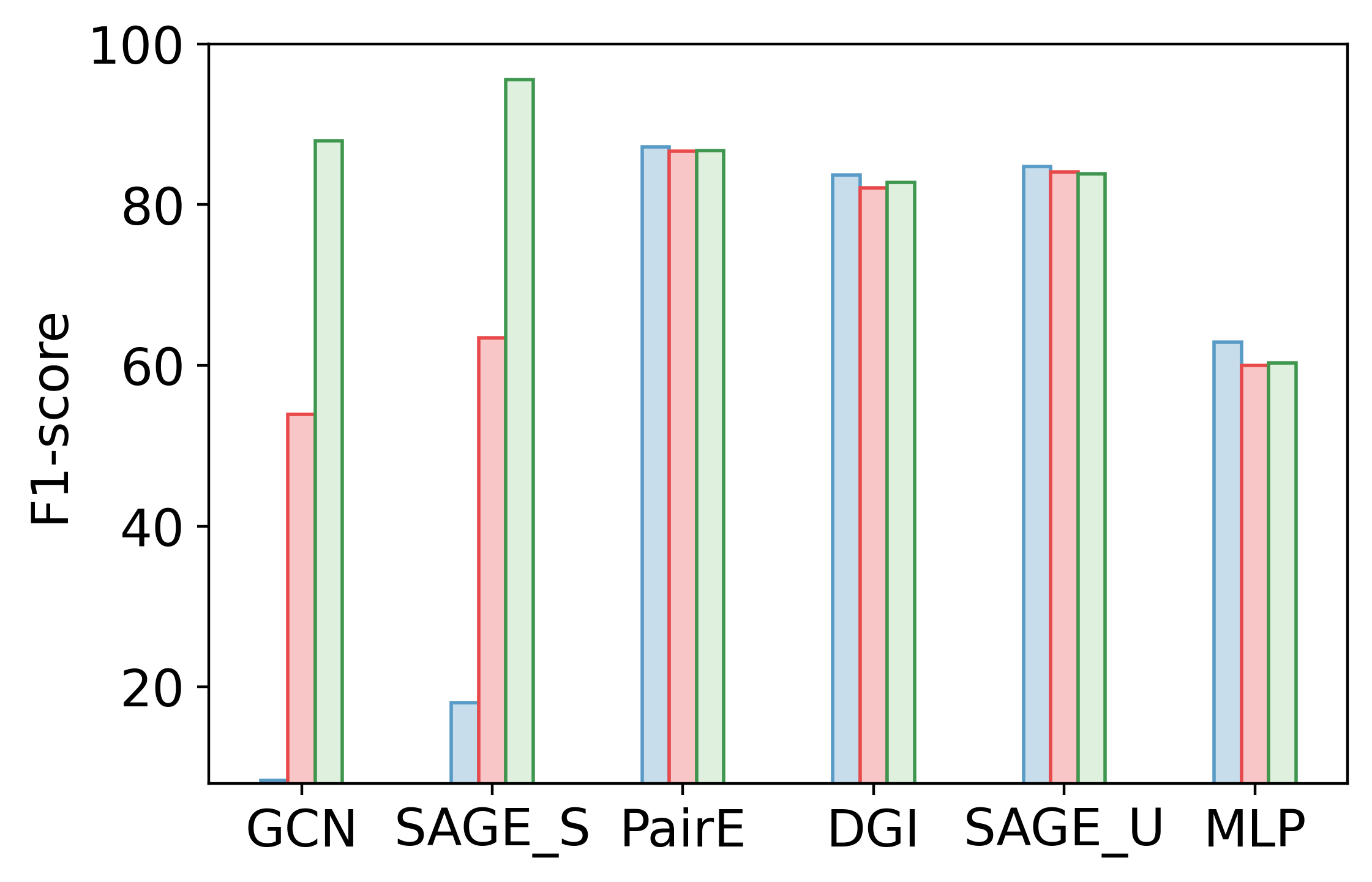}
 \caption{Pubmed}
\end{subfigure}
\begin{subfigure}{0.495\columnwidth}
\centering
\includegraphics[width=0.98\columnwidth]{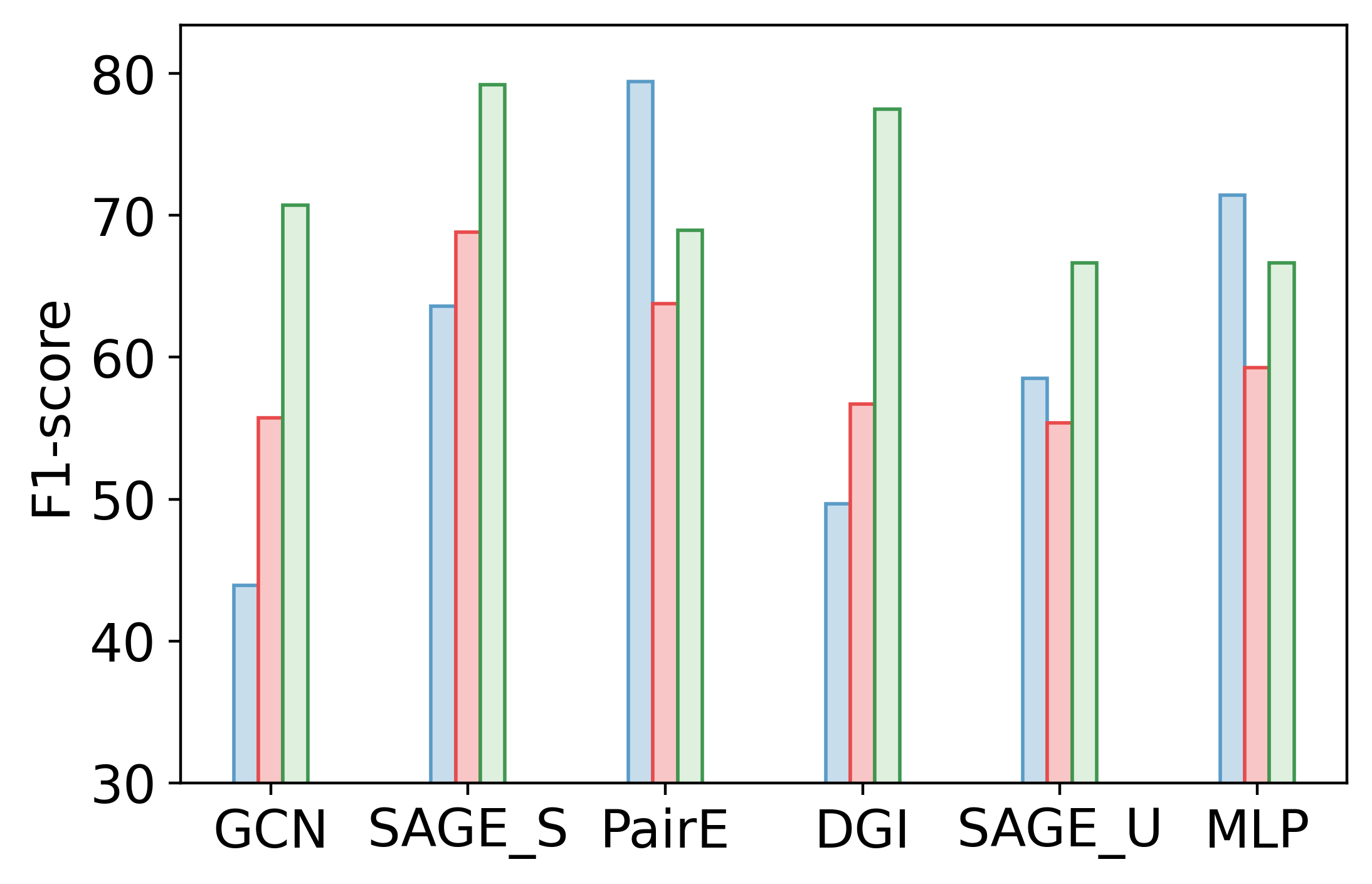}

 \caption{Wisconsin}
\end{subfigure}
\caption{Performance comparisons for node label classification with different local assortativity in Wisconsin and Pubmed}
\label{fig:local_assor}
\end{figure}

First of all, we can intuitively see that PairE has a sound and remarkable effect in the three assortativity levels in the two types of datasets from the Fig.~\ref{fig:local_assor}.

The semi-supervised solutions, e.g., GCN and SAGE-S, have entirely different performances for nodes with different local assortativity: Micro-F1 94.2\% and 93.7\% in Pubmed for nodes with high local assortativity while 7.9\% and 19.7\% for nodes with low local assortativity. It clearly shows that GNN-based solutions can only effectively learn nodes with high assortativity metrics and suffers massive performance deterioration due to the lack of guidance from those long-tail nodes. For GCN and SAGE-S, due to their homophily design, their performance for the node with high local assortativity is much better than nodes with low local assortativity.

However, un-supervised solutions, SAGE-U, DGI, and PairE exhibit much different behavior in  and heterophily networks. In Pubmed, they usually are much less impacted from the node-local assortativity, partially due to their un-supervised embeddings design without the miss-lead from node labels. However, their performance difference becomes noticeable in Wisconsin, and the nodes with high local assortativity generally have much better performance than nodes with low local assortativity. To better illustrate the effect of aggregation, we also prove the result of MLP, which only uses features for classification. For both homophily and heterophily networks, MLP has similar performance towards different types of nodes as it only use node features for classification. 

In short, the prediction performance of a wide range of GNN models display high diversity with respect to the different levels of node assortativity. PairE, in comparison, has stable performance in both Pubmed and Wisconsin and on almost all nodes with different local assortativity. Furthermore, we did observe certain performance degradation for nodes with middle assortativity in Wisconsin. We guess since PairE does not have a pretext to support those nodes with mixed local patterns.

\subsection{Ablation studies}
  In this section, we analyze the effect of different design choices.
    
\begin{figure*}[ht]
\centering

\includegraphics[width=.99\textwidth]{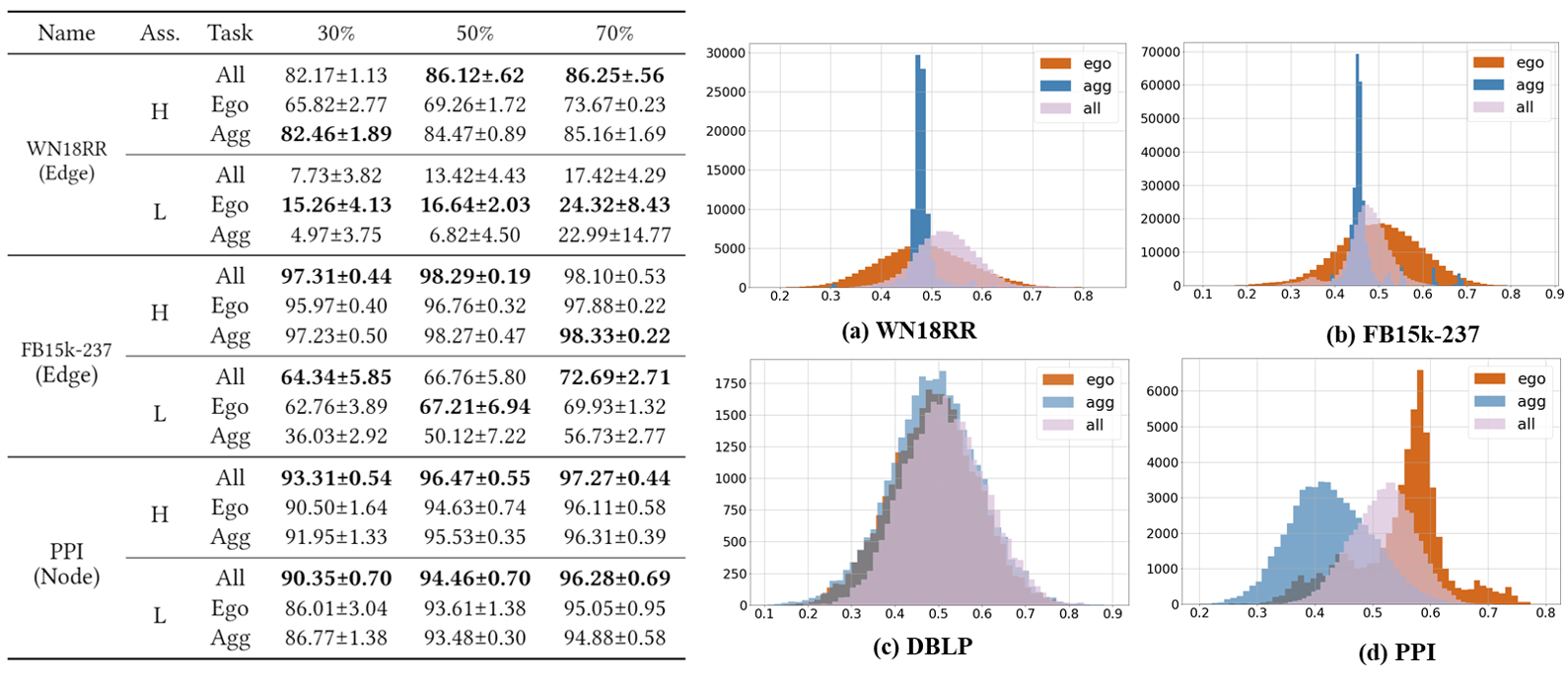}
\caption{Comparison of reconstruction task: performance and embedding distribution. The left table shows the Node/edge classification accuracy(in Micro-F1) with different levels of global assortativity, H/L short for High/Low global assortativity; The $(a)-(d)$ show embedding distributions of their first dimension. Edge embeddings of WN18RR and FB15k-237, and node embeddings of DBLP and PPI.}
\label{fig:assor}
\end{figure*}

\subsubsection{Node v.s. Pair View}
We compare the performance between the pair-view embeddings and the node-view embeddings with the node and edge classification tasks. As can be seen from Tab.~\ref{tab:edge} and Tab.~\ref{tab:node}, compared to  PairE, NodeE has a significant performance difference in two different tasks. 

NodeE suffers significant performance deterioration for the edge classification task, about 68.8\% in FB15K-237 and 75.9\% in WN18RR. The many performances clearly show the importance of pair-view embeddings. It is interesting to see that NodeE achieves excellent performance with only a node's ego and aggregated features for the node classification tasks due to the high assortativity of those networks. In the handwritten dataset Cuneiform, however, node classes are more related to the node positions. Thus, NodeE suffers from the lack of a broader view of the graph topology. In contrast, PairE maintains good performance in all those datasets.

\subsubsection {Ego v.s. Agg. tasks for global (dis)assortative labels}

This section focuses on answering the Q2 from two different perspectives: 1) Do the two reconstruction tasks indeed retain different frequency signals?  2) How will the downstream tasks with different assortativity react to the different types of embeddings?

Please note that the assortativity involved in this experiment is "Global Node/Edge label assortativity".

\noindent\textbf{Performance Comparisons:} The right part of the Fig.~\ref{fig:assor} shows the normalized embedding distribution displays similar generated by different reconstruction tasks or both. As we can see from those figures, all those embeddings follow a Gaussian distribution with different standard deviations $\sigma$. Here, we show only distributions of the first dimension while other dimensions also have similar characteristics. 

In both WN18RR(a) and FB15k-237(b), $\sigma^{ego}$ is much bigger than $\sigma^{agg}$. It verifies our remarks that the ego-task keeps more high-frequency signal, while agg-task, due to its average aggregation, the embeddings from agg-task mainly contains low-frequency signals with small $\sigma$. Compared to WN18RR, the agg-distribution from FB15k-237 is a multimodal distribution with many peaks. We guess that several major distribution patterns of aggregated features can hardly be represented by the edge assortativity metric for overall trends. Fig.~\ref{fig:assor}(c) and Fig.~\ref{fig:assor}(d) show the distributions of the node embeddings generated by the \textit{sum} translator. Fig.~\ref{fig:assor}(c) has three almost identical distribution as DBLP has very high node assortativity$(0.58\sim0.78)$ for almost all labels. 

\begin{figure*}[ht]
\centering
\begin{subfigure}{0.33\textwidth}
\includegraphics[width=\columnwidth]{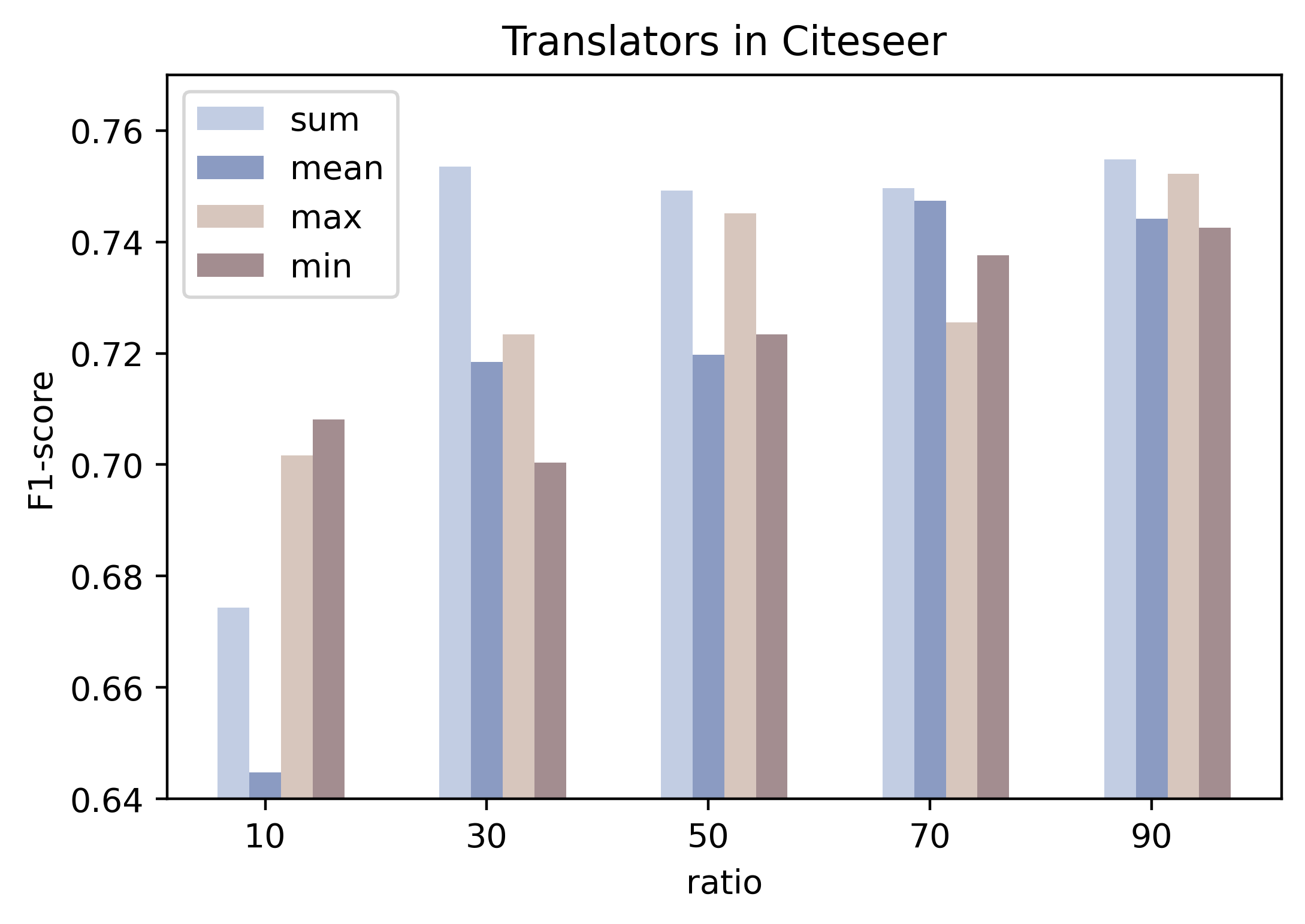}
\caption{Cora}
\end{subfigure}
\begin{subfigure}{0.33\textwidth}
\centering
\includegraphics[width=\columnwidth]{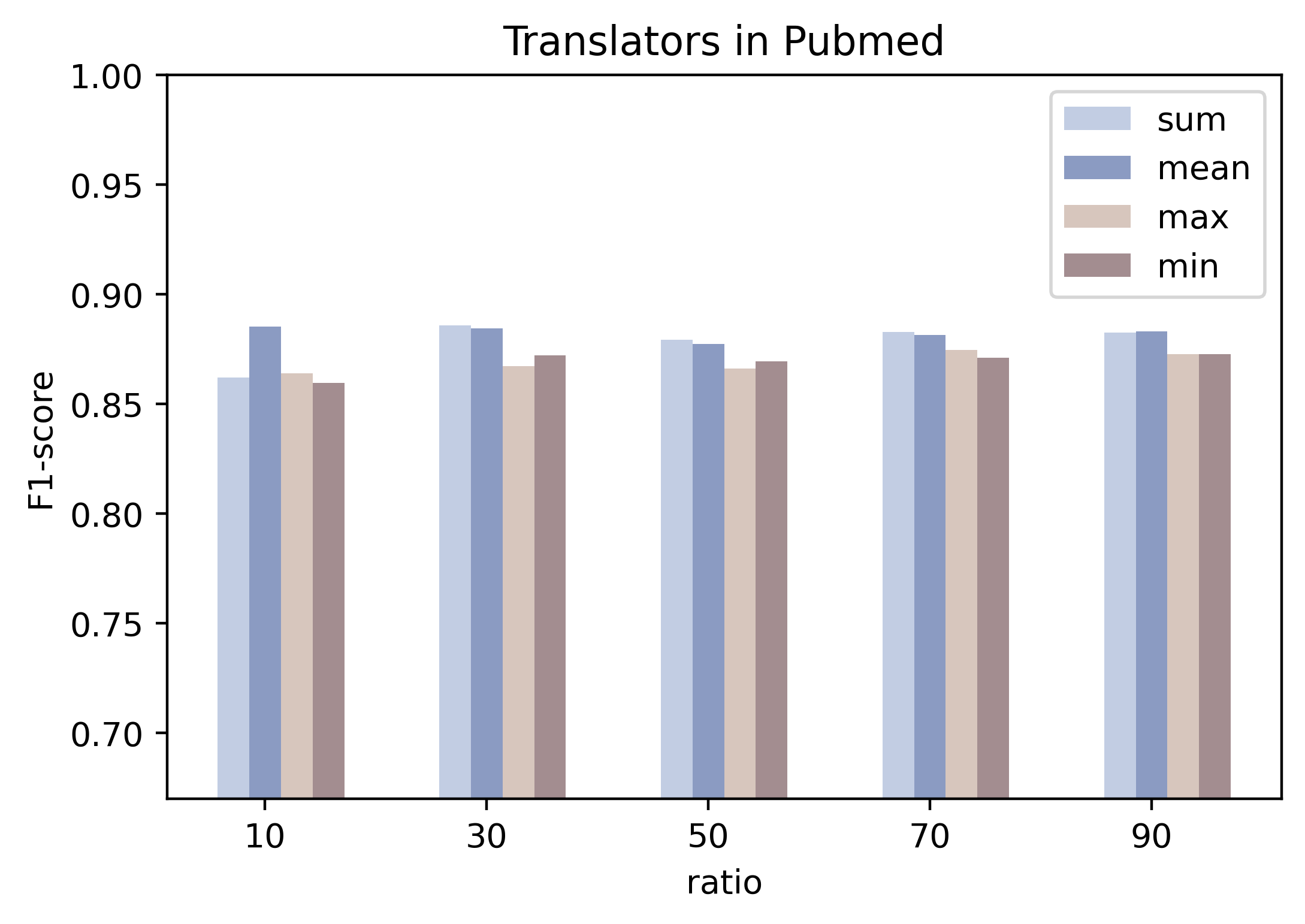}
\caption{PubMed}
\end{subfigure}
\begin{subfigure}{0.33\textwidth}
\includegraphics[width=\columnwidth]{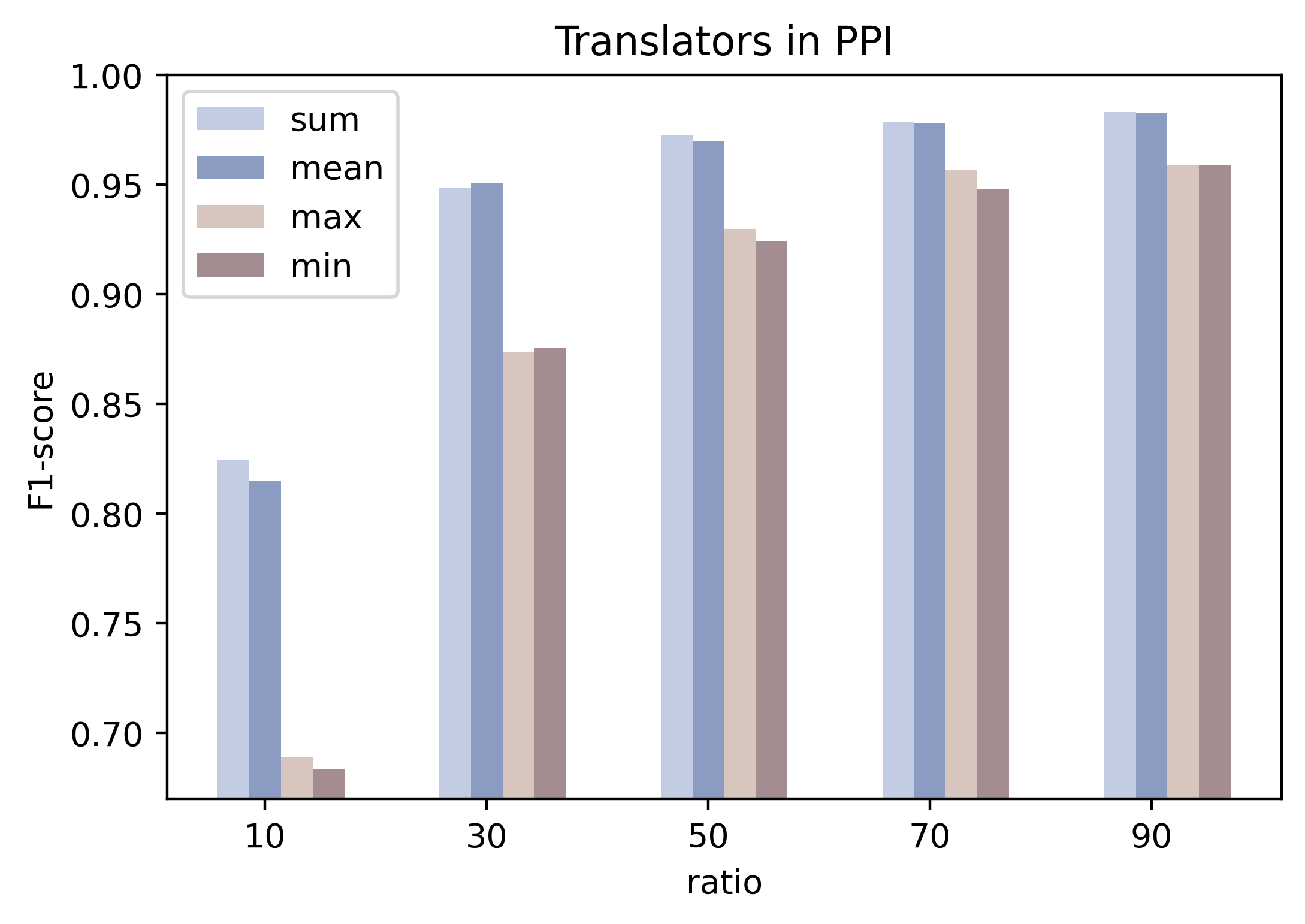}
\caption{PPI}
\end{subfigure}
\centering
\caption{Translators comparisons on Cora, Pubmed, and PPI under different training ratios}
\label{fig:translator}
\end{figure*}

\begin{figure*}[h]
\centering
\begin{subfigure}{0.49\textwidth}
\centering
 \includegraphics[width=0.75\columnwidth]{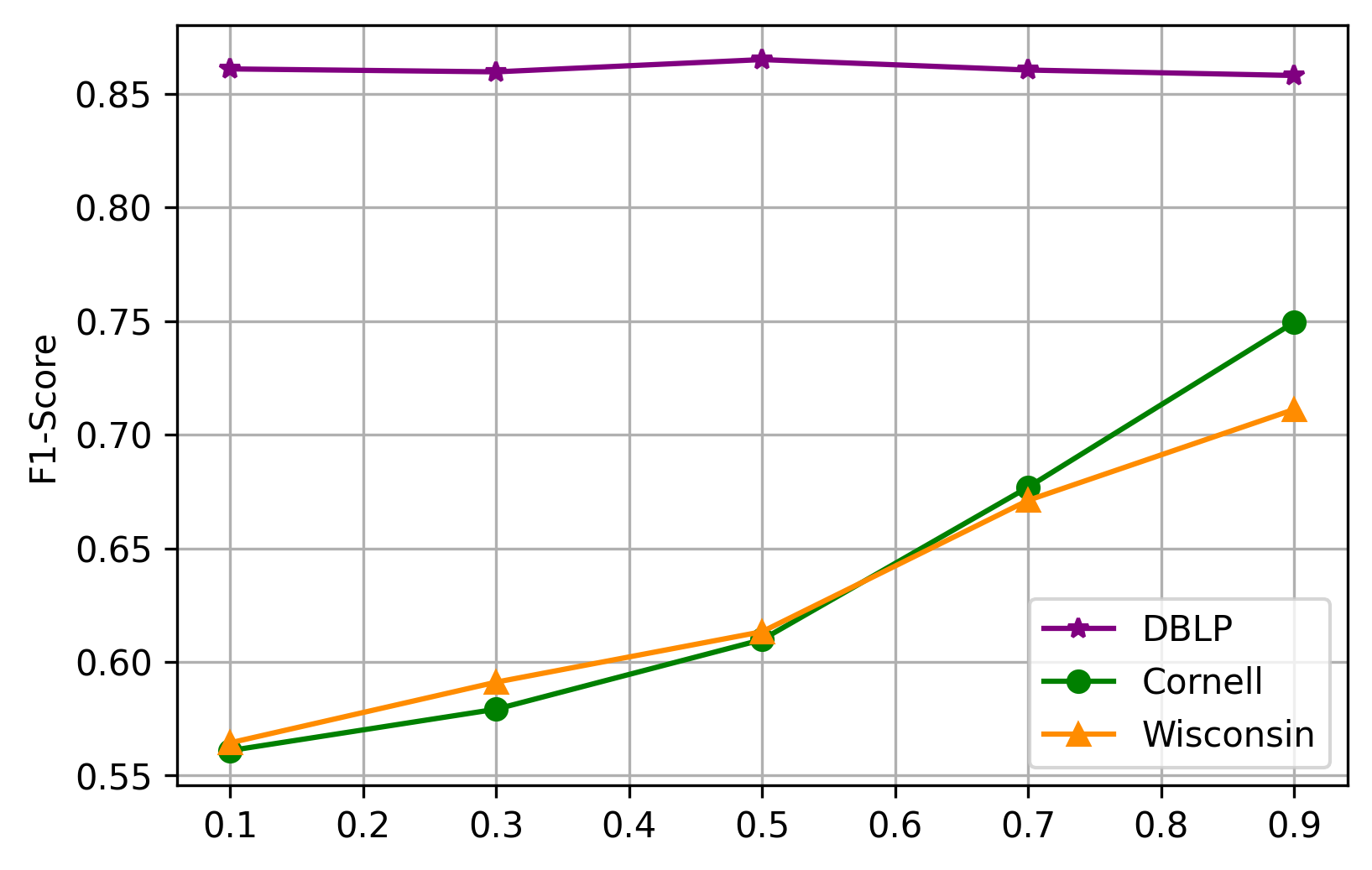}
\caption{Node Classification}
\label{fig:ratio_node}
\end{subfigure}
\begin{subfigure}{0.49\textwidth}
\centering
 \includegraphics[width=0.75\columnwidth]{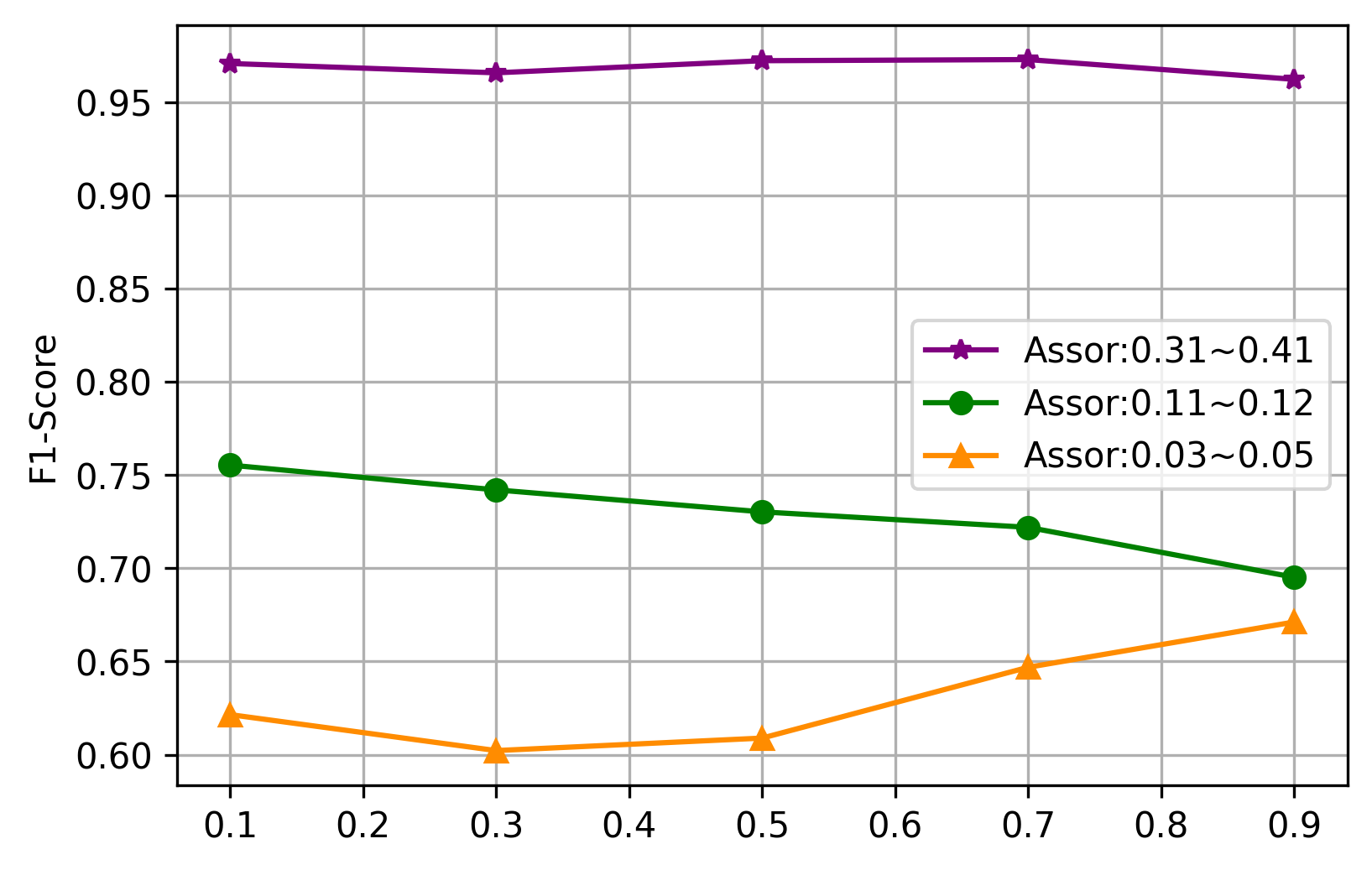}
\caption{Edge Classification}
\label{fig:ration_edge}
\end{subfigure}

\centering
\caption{Node classification(a) and Edge classification(b) results under different weight of two pretext tasks. 0.1 in X axis indicates the weight 0.1 for ego-task and 0.9 for agg-task.}
\label{fig:weight}
\end{figure*}
\noindent\textbf{Impacts towards downstream tasks with different assortativity levels:} To study the impacts of different signals on the tasks with different assortativity, we select the edge/node labels with Top 3 (H) and Lowest 3(L) assortativity for classification. Results are summarized in the left table of Fig.~\ref{fig:assor}. As the four labels of DBLP have similar assortativity, its results are omitted. We can observe that in edge classification, H labels benefit the most from the aggregation task. It has comparable performance with all-tasks, even better in WN18RR, since the aggregation of low-frequency information in the neighborhood provides a robust representation of the edges. However, for low-assortativity tasks, embeddings from ego-task can have much better performance as it contains more edge-specific information towards the edge label. At the same time, the agg-task aggregates much unrelated signals. For low-assortativity tasks: Without the constraints of the modeling object's own information, the graph representation guided by the high-frequency information in the neighborhood will deviate from the ground truth. Therefore, the performance of the agg-task is significantly worse than the ego-task. 

PPI is a typical heterophily network as linked nodes with different types are more likely to connect. Therefore, the embeddings from ego-features and agg-features are pretty different. However, it is also remarkable as the neighbor of a node has displayed certain similarities. Thus, ego-task and agg-task provide similar node classification performance. We guess that there are noticeable patterns in the features of nodes and their surrounding nodes due to the law of chemistry inferred from each other.

\subsubsection {Comparisons of different loss weight between two pretext tasks}
To check the contribution of different pretext tasks and provide an explicit explanation for Q2. We change the ratios of two different losses for the ego-task and agg-task and check the performance variance. As the ratios of two pretext tasks are normalized to one, we only show the weight of the agg-task. Thus, the ratio of 0.1 denotes that the weight for the ego-task is 0.1 and 0.9 for the agg-task. In this case, the ego-task plays a minor role in the back-propagation during training.  Fig.~\ref{fig:weight} illustrates that with the change of data homogeneity, the impacts of ego-task and agg-task towards different downstream tasks. 

Fig.\ref{fig:weight}(a) shows the results for both homophily networks, e.g., DBLP and heterophily networks, Cornell and Wisconsin. For DBLP, the change of ratio might not impact the performance of the node classification tasks as the ego-feature and the aggregated features might have similar distributions. However, for Cornell and Wisconsin, when the weight of the ego-task increases from 0.1 to 0.9, we can find consistent performance improvements as the agg-task might accumulate unrelated information, which might introduce certain noises in the embeddings. 

Fig.\ref{fig:weight}(b) shows its impact on the edge classification task for three groups of representative edge labels with different assortativity levels in FB15k-237. Each group has three edge labels—the three lines in this figure display totally different behavior with the increase of ego ratio. For the labels with high assortativity(0.31$\sim$0.41), the ratio has little impact. The ratio changes also have similar impacts as node classification in Cornell and Wisconsin for the label with very low assortativity. Especially, for the labels with medium assortativity levels, with the increase of the ego ratio, the edge classification task gradually suffers from the higher ratio of ego-task. For those edge labels, the aggregated features might have more stable and robust signals. Thus, for different tasks, the optimal ratio might not be the same. It means the results for the different tasks in Tab. \ref{tab:node} and Tab. \ref{tab:edge} can be further improved as the default ratio of the two tasks is set to (0.5,0.5). 

\subsection{Comparisons of translators}
\label{sec:translator}
This section analyzes and evaluates the performance impacts of different translators in PairE, i.e., sum, mean, max, and min, based on the node classification task.

As can be seen from Fig.~\ref{fig:translator}, the \textit{sum} translator generally achieves the best performance among the four compared translators. It achieves the best performance in almost all the data ratios expect for the 10\% range. The reason that contributes to its good performance, we guess, is that its embedding accumulation operation sums up all pair embeddings starting from the node. To some extent, the "sum" translator keeps certain information on the number of connected edges. The other three types of translators have comparably unstable performance. For instance, the "mean" translator performs well in the Pubmed and PPI dataset while relatively poor in the Cora dataset. However, what operation should be used to best translate pair embeddings to node embeddings with minimal information loss remains a largely unexplored area and could be an exciting direction for future work. 

\subsection{Computation and resource scalability}
Tab. \ref{tab:time} shows the training time of different sizes of graphs in one Nvidia V100-32G for 30 epochs. PairE displays excellent scalability. With the growth of graph scales, the calculation time of PairE increases only linearly. It converges much faster than compared ARVGE, which is also an autoencoder-based solution. For graph achemy\_full with 2 million nodes and 4 million edges, GraphSage, DGI, TransR, H2GCN, and FAGCN suffer the OOM error. Furthermore, PairE is easy to train. It converges typically in about 20 epochs, even in massive graphs. These results answer Q3. The merit is of paramount importance in many real-world applications.
In particular, to support learning beyond the homophily network, H2GCN introduces many designs which bring expensive time and space consumption. Therefore it is challenging to train H2GCN and apply it to big datasets.

  \begin{table}[htbp]
 \centering
  \caption{Trainning time(in seconds) for different scale of graphs, OOM is the out of memory error}
    \begin{tabular}{lrrrr}
    \toprule
    \textbf{Dataset} &  Cuneiform & FB15K-237 & PPI  & alchemy\\
    \midrule
    ARVGE    & 5.95  & 79.84  & 289.06 & 1275.42 \\
    TransR  & 1740 & 19,188 & 55,080 & OOM  \\
    Sage-U    &  48.64  & 188.05  &  647.94 & OOM \\
    DGI & 4.58  & 11.40  & 28.41 &  OOM \\
    H2GCN &5174.48 &OOM & OOM &OOM \\
    FAGCN & 22.23&4064.70 & 1208.4052& OOM\\
    PairE & 5.20& 69.76& 167.44& 519.22\\
    \bottomrule
    \end{tabular}
  \label{tab:time}
\end{table}%

\section{Conclusion}
\label{sec:conclusion}
Real-world graphs usually contain complex types and dimensions of information in both homophily and heterophily networks. To retain both low and high-frequency signals, this paper proposes a novel multi-self-supervised autoencoder called PairE. PairE learns graph embedding from the pair-view, thereby avoiding the loss of high-frequency information represented by edges. The seamless integration of node self-features and nearby structural information is achieved with shared layers in the multi-task autoencoder design to realize the collaborative work of low and high-frequency information. We also provide a solution to translate pair embeddings to node embeddings for node-related graph analysis tasks. Extensive experiments show that PairE outperforms strong baselines in different downstream tasks with significant edges. 

Our experiment results also point out many directions for improvements. For instance, our current solution is unsupervised and task-agnostic. We are working to extend our solution to the semi-supervised message passing-based GNN.  What we want to emphasize is that comparing the performance of a series of models under global assortativity and local assortativity settings, we have observed the limitations of existing GNNs. The most important thing is to discover the interesting and challenging problem of representing graphs from the perspective of local assortativity. This will guide our next steps. 
\subsection{ACKNOWLEDGEMENT}
  This work is supported by the National Science Foundation of China, NO. 61772473 and 62011530148.


  
\bibliographystyle{IEEEtran}
\bibliography{tkde}
\begin{IEEEbiography}[{\includegraphics[width=1in,height=1.25in,clip,keepaspectratio]{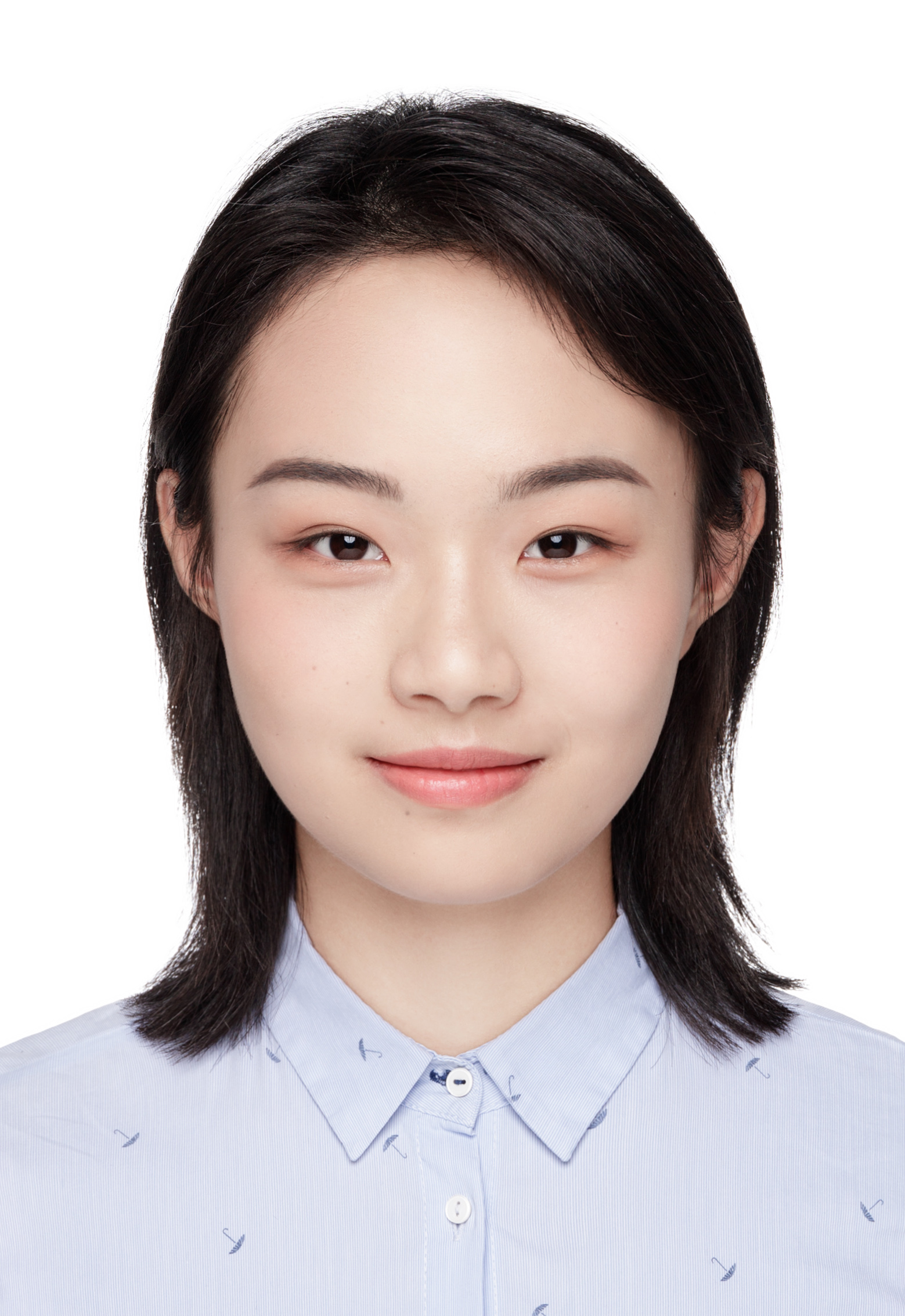}}]{You Li}
is currently working toward the Master degree in the School of Computer Science and Engineering, Central South Of University, China. Her research interests include Graph Representation Learning, and Genetic data retrieval.
\end{IEEEbiography}
\begin{IEEEbiography}[{\includegraphics[width=1in,height=1.25in,clip,keepaspectratio]{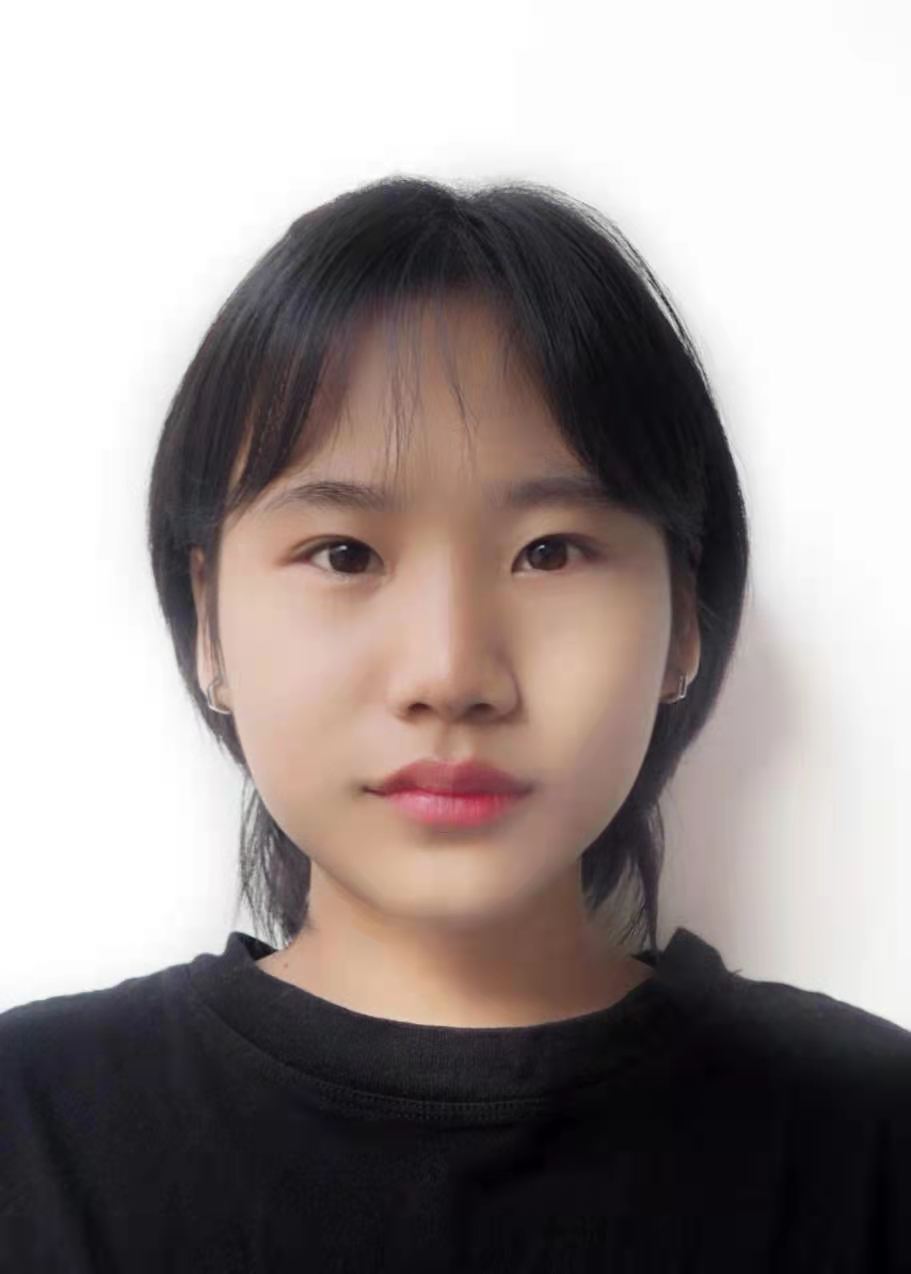}}]{Bei Lin} is currently working toward the Master degree in software engineering at the School of Computer Science, Central South University, Changsha, China. Her research interests include graph representation learning and unsupervised learning.
\end{IEEEbiography}

\begin{IEEEbiography}[{\includegraphics[width=1in,height=1.25in,clip,keepaspectratio]{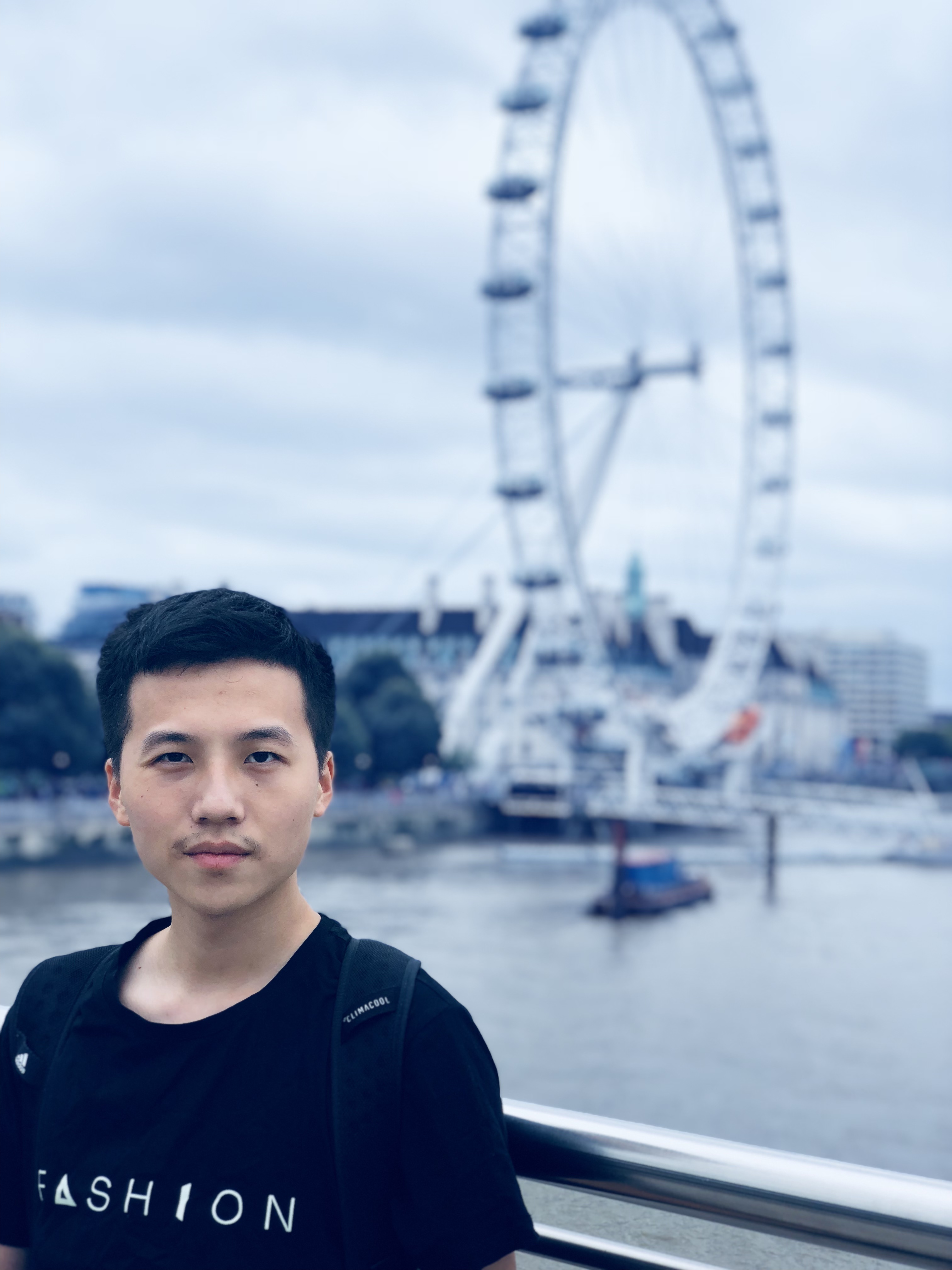}}]{Binli Luo} was a bachelor student in the School of Computer Science and Engineering, Central South Of University, China. His research interests include Graph Representation Learning.
\end{IEEEbiography}
\begin{IEEEbiography}[{\includegraphics[width=1in,height=1.25in,clip,keepaspectratio]{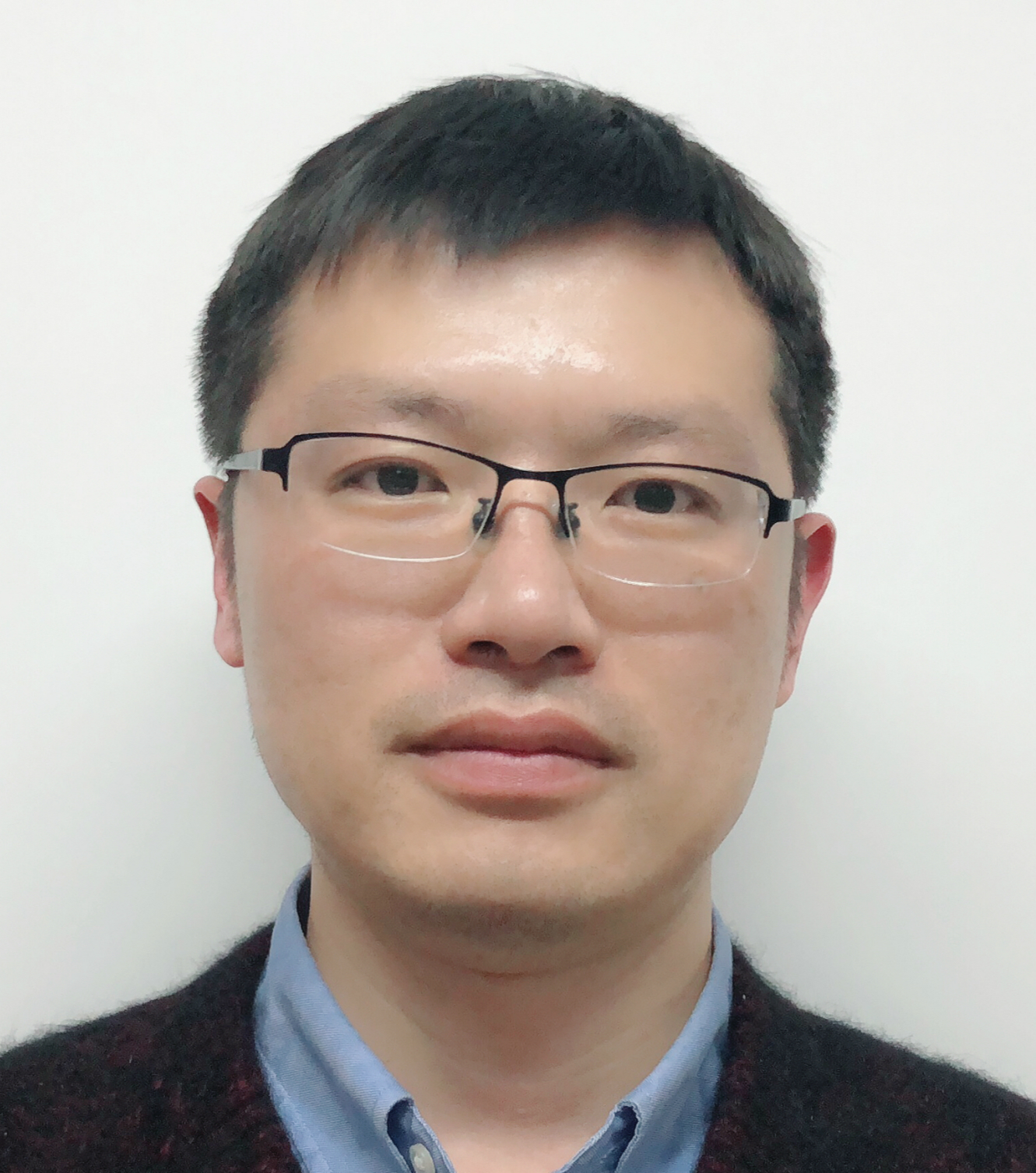}}]{Ning Gui} is associated professor in the School of Computer Science and Engineering, Central South University, China. His research interests include machine learning, representation learning, information retrieval and their applications in the industry. He is a member of the IEEE.
\end{IEEEbiography}

\end{document}